\documentclass[english]{kththesis}

\usepackage{todonotes}

\usepackage[perpage,para,symbol]{footmisc} 

\usepackage{tikz}
\usetikzlibrary{arrows,decorations.pathmorphing,backgrounds,fit,positioning,calc,shapes}
\usepackage{pgfmath}	

\usepackage{rotating}		
\usepackage{array}		
\usepackage{graphicx}	        
\usepackage{float}		
\usepackage{mdwlist}            
\usepackage{setspace}           
\usepackage{listings}		
\usepackage{bytefield}          
\usepackage{tabularx}		
\usepackage{multirow}	        

\usepackage{url}                
\usepackage{hyperref}
\usepackage[all]{hypcap}	

\usepackage{booktabs}
\usepackage{lscape}
\usepackage{pgfplots}
\usepackage{tikz}
\usepackage{ifthen}
\usetikzlibrary{matrix,calc}
\usetikzlibrary{positioning}
\usepackage{adjustbox}

\hypersetup{colorlinks,breaklinks,
            linkcolor=darkblue,urlcolor=darkblue,
            anchorcolor=darkblue,citecolor=darkblue}

\definecolor{darkblue}{rgb}{0.0,0.0,0.3} 
\definecolor{darkred}{rgb}{0.4,0.0,0.0}
\definecolor{red}{rgb}{0.7,0.0,0.0}
\definecolor{lightgrey}{rgb}{0.8,0.8,0.8} 
\definecolor{grey}{rgb}{0.6,0.6,0.6}
\definecolor{darkgrey}{rgb}{0.4,0.4,0.4}
\definecolor{aqua}{rgb}{0.0, 1.0, 1.0}

\usepackage{listings}

\usepackage{csquotes} 

\usepackage{soul}

\usepackage[acronym, section=section, nonumberlist, nomain, nopostdot]{glossaries}
\makeglossaries

\newacronym{NLP}{NLP}{natural language processing}
\newacronym{BERT}{BERT}{Bidirectional Encoder Representations from Transformers}
\newacronym{KBBERT}{KB-BERT}{Bidirectional Encoder Representations from Transformers developed by The National Library of Sweden (sv. Kungliga Biblioteket)}
\newacronym{LSTM}{LSTM}{long short‑term memory}
\newacronym{RNN}{RNN}{recurrent neural network}
\newacronym{NER}{NER}{named entity recognition}

  {\begin{center}
      \selectlanguage{swedish}
      \color{blue}}%
    {\end{center}\selectlanguage{english}
    }
  
\begin{document}
\ifinswedish
    \selectlanguage{swedish}
\else
\selectlanguage{english}
\fi

\title{Period drama}
\subtitle{Punctuation restoration in Swedish through fine-tuned KB-BERT}

\alttitle{Dags att sätta punkt}
\altsubtitle{Återställning av skiljetecken genom finjusterad KB-BERT}

\authorsLastname{Björkman Nilsson}
\authorsFirstname{John}
\email{johnnil@kth.se}
\kthid{u1ezarty}
\authorsSchool{\schoolAcronym{EECS}}

\supervisorAsLastname{Boye}
\supervisorAsFirstname{Johan}
\supervisorAsEmail{jboye@kth.se}
\supervisorAsKTHID{u1ce6y3a}
\supervisorAsSchool{\schoolAcronym{EECS}}
\supervisorAsDepartment{Computer Science}

\examinersLastname{Engwall}
\examinersFirstname{Olov}
\examinersEmail{engwall@kth.se}
\examinersKTHID{u1niocbs}
\examinersSchool{\schoolAcronym{EECS}}
\examinersDepartment{Computer Science}


\date{\today}

\programcode{TMAIM}

\titlepage
\bookinfopage

\frontmatter
\setcounter{page}{1}
\begin{abstract}
  \markboth{\abstractname}{}
Presented here is a method for automatic punctuation restoration in Swedish using a BERT model. The method is based on KB-BERT, a publicly available, neural network language model pre-trained on a Swedish corpus by National Library of Sweden. This model has then been fine-tuned for this specific task using a corpus of government texts. With a lower-case and unpunctuated Swedish text as input, the model is supposed to return a grammatically correct punctuated copy of the text as output. A successful solution to this problem brings benefits for an array of NLP domains, such as speech-to-text and automated text.


Only the punctuation marks period, comma and question marks were considered for the project, due to a lack of data for more rare marks such as semicolon. Additionally, some marks are somewhat interchangeable with the more common, such as exclamation points and periods. Thus, the data set had all exclamation points replaced with periods.

The fine-tuned Swedish BERT model, dubbed prestoBERT, achieved an overall F1-score of 78.9. The proposed model scored similarly to international counterparts, with Hungarian and Chinese models obtaining F1-scores of 82.2 and 75.6 respectively. 

As further comparison, a human evaluation case study was carried out. The human test group achieved an overall F1-score of 81.7, but scored substantially worse than prestoBERT on both period and comma. Inspecting output sentences from the model and humans show satisfactory results, despite the difference in F1-score. The disconnect seems to stem from an unnecessary focus on replicating the exact same punctuation used in the test set, rather than providing any of the number of correct interpretations. If the loss function could be rewritten to reward all grammatically correct outputs, rather than only the one original example, the performance could improve significantly for both prestoBERT and the human group.

\subsection*{Keywords}
NLP, Transformer, punctuation restoration, BERT, KB-BERT, machine learning, neural network

\end{abstract}
\cleardoublepage
\begin{otherlanguage}{swedish}
  \begin{abstract}
    \markboth{\abstractname}{}
Här presenteras en metod för automatisk återinföring av skiljetecken på svenska med hjälp av ett neuralt nätverk i formen av en BERT-modell. Metoden bygger på KB-BERT, en allmänt tillgänglig språkmodell, tränad på ett svensk korpus, av Kungliga Biblioteket. Denna modell har sedan finjusterats för den här specifika uppgiften med hjälp av ett korpus av offentliga texter från landsting och dylikt. Med svensk text utan versaler och skiljetecken som inmatning, ska modellen returnera en kopia av texten där korrekta skiljetecken har placerats ut på rätta platser.
En framgångsrik modell ger fördelar för en rad domäner inom neurolingvistisk programmering, såsom tal-till-texttranskription och automatiserad textgenerering.

Endast skiljetecknen punkt, kommatecken och frågetecken tas i beaktande i projektet på grund av en brist på data för de mer sällsynta skiljetecknen såsom semikolon. Dessutom är vissa skiljetecken någorlunda utbytbara mot de vanligaste tre, såsom utropstecken mot punkt. Således har datasetets alla utropstecken ersatts med punkter.

Den finjusterade svenska BERT-modellen, kallad prestoBERT, fick en övergripande F1-poäng på 78,9. De internationella motsvarande modellerna för ungerska och kinesiska fick en övergripande F1-poäng på 82,2 respektive 75,6. Det tyder på att prestoBERT är på en liknande nivå som toppmoderna motsvarigheter.

Som ytterligare jämförelse genomfördes en fallstudie med mänsklig ut\-värdering. Testgruppen uppnådde en övergripande F1-poäng på 81,7, men presterade betydligt sämre än prestoBERT på både punkt och kommatecken. Inspektion av utdata från modellen och människorna visar tillfredsställande resultat från båda, trots skillnaden i F1-poäng. Skillnaden verkar härstamma från ett onödigt fokus på att replikera exakt samma skiljetecken som används i indatan, snarare än att återge någon av de många korrekta tolkningar som ofta finns. Om loss-funktionen kunde skrivas om för att belöna all grammatiskt korrekt utdata, snarare än bara originalexemplet, skulle prestandan kunna förbättras avsevärt för både prestoBERT såväl som den mänskliga gruppen.

\subsection*{Nyckelord}
NLP, Transformer, återinföring av skiljetecken, BERT, KB-BERT, maskin\-inlärning, neurala nätverk

  \end{abstract}
\end{otherlanguage}
\cleardoublepage

\section*{Acknowledgments }
\markboth{Acknowledgments}{}
I would like to thank the participants of the human evaluation study for taking the time to aid in my research.

\indent Camilla Björkman \\
\indent Jonatan Cerwall \\
\indent Jonas Ekholm \\
\indent Simon Frendel \\
\indent Andreas Furth \\
\indent Anders Nilsson \\
\indent Max Nilsson \\
\indent Rasmus Nilsson \\
\indent Thomas Nilsson \\
\indent Hugo Olsson \\
\indent Elias Soprani \\
\indent Gisela Soprani \\
\indent Eva Wahlfridsson \\
\indent Dag Wahlfridsson \\
\indent Linnéa Wahlfridsson \\
\indent Anna Zettersten \\

I am glad that many of you had fun doing the test. Without you, this report would not have been as interesting!

Additionally, I want to thank my supervisor Johan Boye for taking a lot of time reading this report and coming up with ways to improve it. I will keep this praise brief and non-hyperbolic, just like he taught me.

\acknowlegmentssignature

\fancypagestyle{plain}{}
\renewcommand{\chaptermark}[1]{ \markboth{#1}{}} 
\tableofcontents
  \markboth{\contentsname}{}

\cleardoublepage
\listoffigures

\cleardoublepage

\listoftables
\cleardoublepage
\printglossary[type=\acronymtype, title={List of acronyms and abbreviations}]
\label{pg:lastPageofPreface}
\mainmatter

\renewcommand{\chaptermark}[1]{\markboth{#1}{}}
\chapter{Introduction}
\label{ch:introduction}
\section{Background}
\label{sec:background}

Human language is something we all engage in every day, whether it is through conversation or ordering home delivery through an app. Even though our languages are all unique in their own way, they are all complex and hard to understand without having used them often and for a long time, preferably since birth. Their complexity carries over when trying to make computers understand them. In our increasingly technologized world, the desire for computer-made human language arises for a myriad of different applications. Examples include customer service chatbots, email spam filters, voice controlled devices, search engines, text analytics such as summary generation, and translation. This field of study is known as \acrfull{NLP}. This study investigates the task of punctuation restoration, i.e. the act of reintroducing punctuation such as periods, commas and question marks into pieces of text. The need for this solution has come up in a larger-scale research project at KTH. The goal of that project is to automatically generate question-answer pairs for the purpose of reading comprehension tests.
For example, the input

\vspace{5mm}

\fbox{\begin{minipage}{25em}
    A financial controller's main task is to analyze the past and present situation in economic figures. Being able to read results and goal fulfillment and then report to management and to others in the organization is the most important thing.
\end{minipage}}

\vspace{5mm}

\noindent{the question-answer pair}

\vspace{5mm}

\fbox{\begin{minipage}{25em}
Q: What is the main task of a financial controller? \newline
A: To analyze the past and present situation in economic figures
\end{minipage}}
\vspace{5mm}

\noindent{could be the generated output, albeit with all text in Swedish.}

This task has been empirically found to be easier if the input text is all lowercase characters with no punctuation. This means that there is a need for a separate program to reintroduce correct casing (capitalization) and punctuation in the question and answer pair. The task of punctuation restoration, but not capitalization, is the focus of this report. The method presented here should take an input such as

\vspace{5mm}

\fbox{\begin{minipage}{25em}
    can you fly this plane and land it
    surely you can't be serious
    i am serious and don't call me shirley
\end{minipage}}

\vspace{5mm}

\noindent{and then}

\vspace{5mm}

\fbox{\begin{minipage}{25em}
    can you fly this plane and land it?
    surely you can't be serious?
    i am serious, and don't call me shirley.
\end{minipage}}

\vspace{5mm}

\noindent{could be the generated output, albeit with all text in Swedish.}

\vspace{5mm}


Similar projects in other languages have shown promising results by using the BERT type of language model for punctuation restoration. However, performance differences suggest that it is almost necessary to have a language model tailor-made for the specific language that the model will be used on, rather than having a multi-language solution \cite{hungarian, adversarial}. Training this sort of model requires substantial amounts of data \cite{devlin2019bert}.
The National Library of Sweden naturally have a large corpus of written material in the Swedish language. The organisation realized the many uses a Swedish BERT could have, ranging from practical examples like a sort of spam filter for fake news to intellectually interesting examples such as tracking a famous author's stylistic influences in the language over decades \cite{kb-nyheter}. In early 2020, their data science department released three different BERT models that outperformed Google-developed multilingual models \cite{kb-bert}.

Since KB-BERT is still relatively new, there is still potential for solving many different problems, often inspired by what has been done in other languages. One such task is that of punctuation restoration that has already be done successfully abroad \cite{hungarian, adversarial}. Restoring punctuation is an important task in a variety of NLP endeavours. Most state-of-the-art NLP models are pre-trained on punctuated data, meaning that performance can drop over 10\% when the models are deployed on unpunctuated text \cite{alam}. Additionally, the rise of speech-controlled devices has increased the amount of unpunctuated text (in this case, transcriptions) in need of punctuation restoration \cite{bangladesh_presto}.

\section{Research Question}
How does a fine-tuned KB-BERT model perform at punctuation restoration compared to humans in Swedish and international BERT models in their respective languages?

\section{Contributions}
The contributions of this report are two-fold:

\begin{enumerate}
    \item a fine-tuned Swedish BERT model for punctuation restoration available at huggingface.co/Johnnil/prestoBERT and
    \item an investigation into how difficult the problem is through a human evaluation baseline study.
\end{enumerate}

\section{Delimitations}
Similar projects with different languages have added complex additions upon the "simple" pre-trained BERT and fine-tuned layer architecture. 
The goal of this project is not to develop a state-of-the-art model that rivals those of international research teams with funding, but rather explore the possibilities of applying the same core approach of a language-specific BERT model to Swedish. Due to time restraints, the experimental additions have thus been left out.

A data set of size 2.7 megabytes has been provided by the supervisor of this project. Gathering more data is doable, but involves scraping data sets and cleaning that data, which is time consuming. There is also some value in trying a homogeneous data set and seeing how it generalizes.

\cleardoublepage

\chapter{Background}
\section{Statistical and neural NLP}
In this section it will become clear that neural networks is the most suitable method for solving the problem of this thesis.

NLP as a field has existed since the 1950s. In its infancy, NLP methods centered around learning a set of rules, i.e. symbolic NLP, for the purpose of translating between two arbitrary languages. Human engineers provided the computer with dictionaries and grammatical rules so that it could properly conjugate output words and adhere to word order conventions. In other words, defining set rules like the ones taught to people in school. It requires significant human efforts to detail every possible rule imaginable in code \cite{hutchins2005history, progress_neural_NLP}. The symbolic approach, dating back to the pre-internet era, does not utilize the large amount of data available online. 

The 1990s came with a paradigm shift, statistical NLP. The new "example-based" translation showed great potential. Compared to the earlier symbolic approach of explicitly formulating rules, statistical NLP provided the computer with a data set of examples. Computers could successfully use correct grammar when generating their own sentences through the implicit rules in these large corpora and the increasing availability of them meant that statistical NLP became the new norm \cite{hutchins2005history}. Statistical NLP makes use of the aforementioned data and offers more robustness, higher performance and speed \cite{why_statistical}. This eventually made symbolic NLP obsolete, at least in the context of machine translation \cite{hutchins2005history}. Statistical NLP eventually evolved into including neural networks to automate the process of learning from data \cite{why_statistical}. Looking at the top performing models, i.e. state of the art, for the problem of punctuation restoration in any language, every single entry found is a BERT-based neural network \cite{devlin2019bert, hungarian, adversarial}.

The next shift in NLP methodology came with the advent of neural networks in the 2010s. While still a statistical methodology, neural networks offer less interpretability. Despite this flaw, their increased performance made them the only viable choice going forward. Neural networks in NLP work with something called "word embeddings". The core idea is not unique either to neural networks or word vector, but actually dates back to the 1950s when J.R. Firth (1957) suggested that we "\textit{... shall know a word by the company it keeps}" \cite{firth1957synopsis}.

\subsection{Word embeddings}\label{sec:word_embeddings}
A stand-out feature of statistical NLP is the concept of word embeddings. The neural network needs to develop semantic knowledge of the language. To this end, word embeddings is an abstract representation of a word, in practice a long vector of numbers. If two vectors are similar to each other, it means that their respective words are related. In the word embedding generating program \textit{Word2vec}, the resulting word embeddings also enables the following function:

\vspace{5mm}

\fbox{\begin{minipage}{25em}
    Input: vector(Paris) - vector(France) + vector(Italy)
    
    Output: vector(Rome)
\end{minipage}}
\vspace{5mm}

\noindent This is because word embeddings appear to capture the relationship between words, such as "capital of" between Paris and France and Rome and Italy respectively. This makes the above input much like a rebus: "capital of \st{France} Italy" \cite{mikolov2013efficient}.

The concept of word embeddings are quite intuitive.
For example, if the words "chair" and "couch" tend to come up in similar contexts, such as "I sat on the ...", logical reasoning motivates treating these two words as similar \cite{attention
}.
This is sort of similar to building the understanding of what a chair is from observing the sentences "I sat on the chair by the kitchen table" (location) and "I could only carry two chairs at the same time" (unwieldiness) to build an understanding of what the word means. Imagine reading upwards of hundreds of thousands of contexts for each word and it is intuitive to understand how a neural network develops semantic knowledge solely from statistics, without predefined rules. The key is to have many different examples. 

Word embeddings are numeric vectors that attempt to capture this semantic knowledge in mathematical form. They are \textit{embedded} in the vector space. The motivation for this approach is the previously mentioned idea that "you shall know a word by the company it keeps" (J.R. Firth, 1957) \cite{firth1957synopsis}, which is another phrasing of the argument in the previous paragraph. There is evidence that words can be similar in more than one way, which motivates the use of multidimensional vectors \cite{word_similarity}.

\begin{figure}[!ht]
  \begin{center}
    \includegraphics[width=0.8\textwidth]{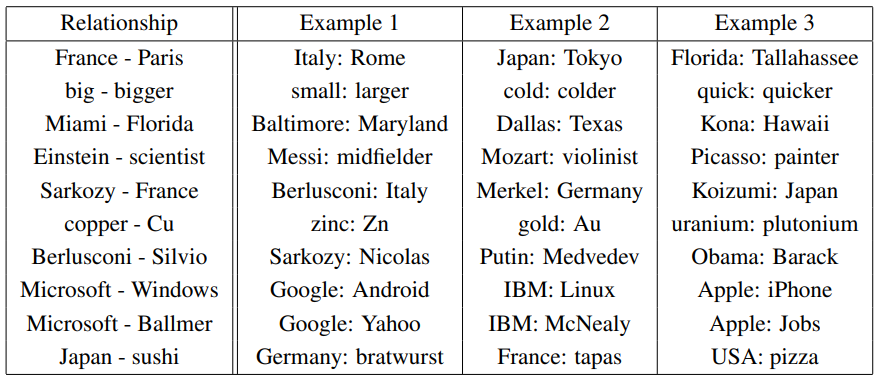}
  \end{center}
  \caption{Examples of the word pair relationships, using the best word vectors from Skipgram model trained on 783M words with 300 dimensionality (from \cite{mikolov2013efficient})}
  \label{word_pair_relationships}
\end{figure}

In the Word2vec method of generating word embeddings, the resulting embeddings capture word similarity in more ways than syntactic, as shown in the capital city example earlier, where Paris and Rome were identified to both be capital cities of their respective countries. The paper introducing Word2vec also gives the king and queen example, showing that the program can deduce the gender associated to gender-specific titles \cite{mikolov2013efficient}, possibly by observing gender pronouns appearing close to the titles as well as the words "woman" and "man". More examples of word pair relationships can be seen in figure \ref{word_pair_relationships}.

One shortcoming of the Word2vec method is that the generated embeddings are context-independent, meaning that each word in the dictionary only has a single embedding where all interpretations of the word are included \cite{context_dependence}. For some words, such as "Nice", which can mean both an expression of positivity and a city in France, this is problematic. The non-city definition of "nice" is a common word, whereas the city is rarer. This means that the word embedding for the city of Nice is farther away from the cluster of other French cities like Paris and Bordeaux.

\section{Recurrent neural network}\label{sec:RNN}
In order to process sentences of any kind, some sort of memory is required in order to keep track of the entirety of the sentence. For example, in the incomplete sentence "The drill fell into the hole that", the next word is rather likely to be a pronoun. When choosing an appropriate pronoun, the network must remember the word related to the pronoun, in this case the subject of the sentence; drill. A drill is an object most often not referred to as "he" or "she" but rather "it". If the network can not look back far enough, it might not realize that a pronoun is likely or choose the wrong one. Traditional feed-forward neural networks do not support a short-term working memory.

The \Acrlong{RNN} (RNN) type of neural network performs the same task over each instance of an input sequence. The RNN maintains a hidden state that is updated in each time step, hence the term "recurrent". A hidden state or vector is essentially information that is neither connected to the input nor the output directly. Thus, they are not observable during normal use of the model, in other words they are hidden from view. Hidden parts of neural networks can be difficult to interpret, as, unlike input and output, they are not in a form that is meant for humans to understand.

The hidden state functions as a sort of memory of earlier computations, which makes the RNN unique compared to other neural network architectures. The next time step is dependent on the computation and result of all the previous time steps, using information about the context carried in the hidden state. Young et al. \cite{rnn} explains why this is favorable for NLP tasks with the example "hot dog" where the word "hot" before "dog" significantly modifies the semantic meaning of "dog". Remembering that "hot" occurred earlier in the sentence is key. Taking it a step further, to predict the missing word in the sentence
\begin{equation}
    \textnormal{"Earlier, I ate a hot \_\_\_"}
    \label{hot_dog_stand}
\end{equation}

remembering and taking "ate" into consideration to boost food-related predictions could be key in correctly guessing "dog".

\begin{figure}[!ht]
  \begin{center}
    \includegraphics[width=0.8\textwidth]{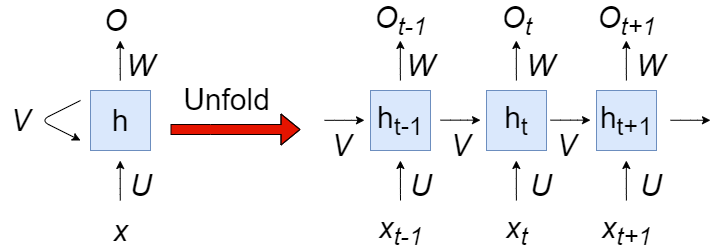}
  \end{center}
  \caption{Structure of a simple RNN (based on \cite{rnn_bild})}
  \label{rnn_schema}
\end{figure}

Firstly, the RNN architecture supports variable length input, meaning that anything from a single word to large documents \cite{variable_length_rnn}. While flexible, the RNN does not handle long-term dependencies well, according to empirical results. In other words, it forgets too quickly \cite{rnn}.

Secondly, RNNs suffer from \textit{the vanishing gradient problem}. When updating network weights during the learning phase, it is done by sending gradients as instruction for which direction the weights should be updated. For individual weights, this is a one-dimensional value of how much it should increase or decrease. These gradients are calculated from the error function where larger errors mean a larger increase or decrease to correct. These calculations are propagated, from the output in the front, backwards in the network. The network decision process is akin to a sort of decision-making baton-pass. The earlier time step computations lay the foundation of information that the later computations make their decisions upon. This means that any time step calculation in the network could be blamed for any error, but it is significantly easier to determine blame close to the output, where it is clear exactly which prediction was correct (and should have been guessed with higher probability) and which is incorrect (and should have less probability). This leads to the vanishing gradient problem, meaning that as gradients are updated using back-propagation, the gradients shrink into close-to-zero values towards the earlier time steps. As a consequence, earlier computations from the earlier time steps lead to significantly less training for the network compared to later time steps \cite{lstm}. In the RNN architecture, there is only a single neuron, but the structure can be "unfolded" to show the hidden state $h$ in each time step, as seen in figure \ref{rnn_schema}, which function similar to layers of multiple neurons in a more general artificial neural network in terms of backpropagation.

Lastly, the repeated updating of the same hidden state makes RNNs inherently sequential; state $h_{t}$ at time $t$ is dependent on states $\{h_{0}, ..., h_{t-1}\}$. Recent developments have, through factorization tricks and conditional computation, offered increased computation efficiency, but it is impossible to escape the sequential nature that precludes true parallelization \cite{attention}. This makes RNNs significantly less able to take advantage of modern hardware relying on multicore computation, such as GPUs and TPUs \cite{uszkoreit}.

\subsection*{LSTM}
Introduced in the late 90s, the \acrfull{LSTM} architecture was introduced as a solution to the \acrshort{RNN}'s shortcomings. It also solves problems \acrshort{RNN}s could not previously do. The name stems from the supposed fix of making sure that the short-term memory does not decay at the same high rate, essentially making it \textit{longer} short-term memory 
\cite{lstm}.

\section{Encoder-decoder}
The next step in NLP came in 2014 with the introduction of encoder-decoder networks. By combining two separate RNNs, one encoder and one decoder, that share a 
vector representation, the concept of word embeddings for full sentences are incorporated \cite{encoder_decoder}. For example, in translation tasks between two languages, one language is encoded into an abstract context in the form of a hidden vector. The sentence "I am hungry" is encoded into a mathematical representation, thought to be the semantic concepts involved, such as a subject describing itself with an adjective. There is no way to verify whether this is true, but it is an intuitive theory of what the vector contains. This sentence embedding is then decoded into potentially any other language, even if it involves fewer or more words to convey the same semantic meaning. Once again, the approach is weak when dealing with longer sentences \cite{attention_intro}.

\begin{figure}[!ht]
  \begin{center}
    \includegraphics[width=0.6\textwidth]{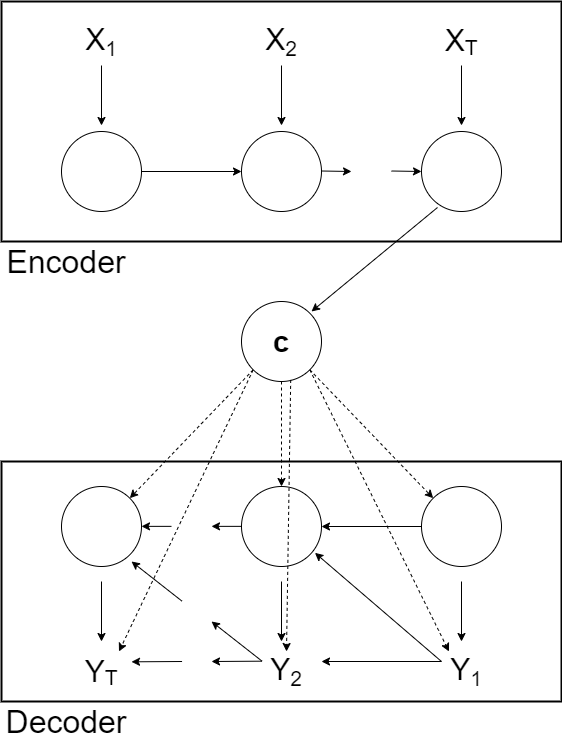}
  \end{center}
  \caption{RNN encoder-decoder transforming input sequence $x_{1}, ..., x_{T}$ into output $y_{1}, ..., y_{T'}$. Dashed arrows indicate conditioned dependencies on the context vector \textbf{c}.
  From \cite{encoder_decoder}.}
  \label{fig:encoder_decoder}
\end{figure}

The encoder-decoder network combines the concepts of word embeddings with RNNs. An encoder-decoder consists of two stacked RNNs: one encoder and one decoder. Visualized in \ref{fig:encoder_decoder}, the two networks join in node $c$, the last hidden state of the encoder RNN, also known as the context vector. This is a representation of the input sequence, i.e. an abstraction of the semantic meaning of the sentence. Much like the regular RNN, the same flaws affect the performance of the encoder-decoder \cite{rnn}. For example, only the last hidden layer from the encoder is passed on to the decoder, which means that the earlier symbols in the input, such as $x_{1}$, has significantly less effect on $c$ than the later ones, such as $x_{T}$. 

\subsubsection*{Encoder}
The encoder encodes a variable-length input sequence to fixed-length context vector $c$. This vector, much like word embeddings, carries a high-level representation of the sentence, called the \textit{context}, of the input sentence. The encoder is trained jointly with the decoder, for example by translating between two languages. When introduced by Cho et al., they trained the two RNNs jointly on maximizing the conditional probability of a French sentence given its English counterpart, such as $p(bonjour | good morning)$ or generally $y_{1}, ..., y_{T'} | x_{1}, ..., x_{T}$ \cite{encoder_decoder}.

\subsubsection*{Decoder}
Connected to the encoder through its final hidden layer, the decoder is tasked with decoding the abstract representation. It is trained on predicting the next symbol $y_{t}$ given previous symbol $y_{t-1}$ and current hidden state $h_{t}$ \cite{encoder_decoder}.

\section{Attention}\label{sec:attention}
The attention mechanism is about measuring what is relevant of what is not. Have you read everything in this report up until now? If not, you might be using a form of \textit{attention} by looking up this particular section to find out more about attention. This section is, after all, more relevant for learning about attention than any other section in the report, so it makes sense to read this rather than \ref{sec:word_embeddings} Word embeddings.

This encoder-decoder shortcoming of not handling longer sentences well is addressed in the same year by the concept of "attention", that successfully circumvent the bottle neck of the fixed-length hidden vector by allowing the network to search for relevant information pertaining to specific tasks. This means that when trying to decode a certain part of the embedding, the network can look at the attention of it. For example, in the machine translation between languages of different language trees, the word order is often different. This is not a problem if the network can see which words relate to which and to which degree \cite{attention_intro}. 

In section \ref{sec:RNN}, we looked at sentence \ref{hot_dog_stand} as an example of what RNNs, and by extension encoder-decoder networks, struggle with. If a human were to guess the blank word, "hot" and "ate" is likely sufficient to make "dog" a common guess. Even just "hot" might make "dog" the single most likely word. If we only showed the words "I ate" or "Earlier", it would be much more difficult to correctly guess dog. This is because not all words are equally important in regards to the missing word. That is the intuition behind the attention mechanism.

Instead of letting a fixed-length vector represent the whole sentence, like the RNN, allowing the decoder to look back upon the input sequence could help it make better predictions \cite{attention_intro, rnn}. For example, in the case of missing word prediction, if a country name (e.g. "France") is present in the input sentence and it is likely that a missing word should be a language (e.g. "spoke fluent \_\_\_"), connecting mentioned countries and their language(s) is an easy way to boost performance in the average case. 
In addition to the context vector, the attention mechanism in an encoder-decoder network saves a copy of every hidden state $h_{t}$ to send to the decoder. The input sequence containing words $x_{0}, ..., x_{T}$ are each annotated with annotations (encoder hidden states) $h_{0}, ..., h_{T}$, containing information regarding the entire input sequence with a focus on the words surrounding the corresponding $x_{i}$-th word. Then, for each pair of decoder hidden state $s$ and annotation $h$, an attention score (sometimes called energy) $e^{\langle s, h\rangle}$ is computed. This score tells the decoder how important each word of the input is when generating the next word \cite{attention_intro}. For example, in the hot dog example, "ate" and "hot" would have high values for $e$ whereas "Earlier" would have a low score.

\begin{figure}[!ht]
  \begin{center}
    \includegraphics[width=0.95\textwidth]{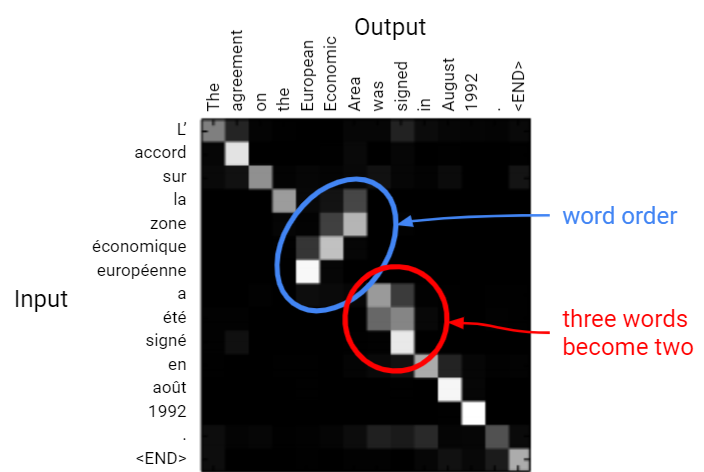}
  \end{center}
  \caption{Word alignment matrix for English-to-French example. White indicates high attention, black indicates no attention. Based on \cite{attention_intro}.}
  \label{fig:attention_illustrated}
\end{figure}

The $e$-scores are then put through the Softmax function, a non-linear scaling to put all scores in the range $[0, 1]$ with $\sum{softmax(e)} = 1$, i.e. turning them into probabilities $\alpha^{\langle s, h\rangle} = softmax(e^{\langle s, h\rangle})$.
The attention weight given to each annotation is equal to its $\alpha$-value, meaning that they each get a proportional share of the attention.
As a side-effect of Softmax, the differences between small and large values of $e^{\langle s, h\rangle}$ are increased, meaning that semi-relevant words end up mattering significantly less that highly relevant words. This effect can be visualized in figure \ref{fig:attention_illustrated} where most word relationships have virtually no attention between them, making the image have substantial contrast rather than a softer grey.
This enables "hard alignment", also referred to as "hard attention", where the attention scores $\alpha^{\langle s, h\rangle}$ are equate how much attention is given in that computation. Due to Softmax, the gap between values tends to be larger. This makes hard attention focus on a select few items. Soft alignment, also known as "soft attention", is when each $s$-value gets a share of the attention equal to its value, essentially spreading the attention wider. Which one is best suited depends on the task, but in this case, soft attention is used \cite{attention_intro}.

From this figure (\ref{fig:attention_illustrated}), we can see how attention aids beyond word-for-word translation (visualized as a diagonal line in the figure). Firstly, the English translation of "a été signé" is only the two words "was signed", which means that a simple word-for-word approach is insufficient. Through attention, the word "was" has been generated by looking at both "a" and "été". 
Secondly, the word order for "zone économique européenne" and "European Economic Area" is different. 
Attention manages this by looking at all words and deciding that "européenne" is the most relevant word to look at when generating the next word after "The agreement on the".

\subsection{Self-attention}\label{sec:self_attention}
\begin{figure}[!ht]
  \begin{center}
    \includegraphics[width=0.8\textwidth]{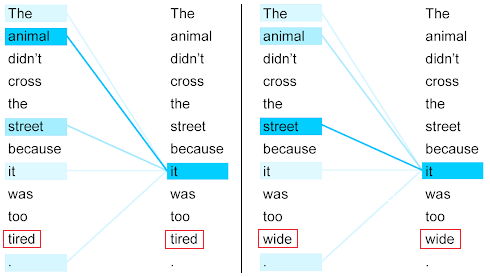}
  \end{center}
  \caption{Attention example visualized. Blue indicates high attention, white indicates no attention. The red boxes indicate which word is different between the two examples. Based on \cite{uszkoreit}.}
  \label{fig:cross_the_street}
\end{figure}

Another strength of attention is only found in a specific version called \textit{self-attention}. It has the ability to use context from the entire input sequence when encoding the words. Rather than regular attention, self-attention models the attention between different tokens within a sequence, rather than attention between two sequences. In other words, self-attention calculates attention between words in the input rather than between the $i$-th input and $k$-th output word. Compared to regular attention, self-attention better estimates the modified word in the case of ambiguous modifiers not only backwards in the sentence, but also forwards, i.e. bidirectionally. This is useful in the example shown in figure \ref{fig:cross_the_street}. The word "it" can refer either to the animal or to the street. Based on real-world experience, one can deduce that it is unlikely that an animal would be "too wide" to cross a street or that an inanimate object being personified as "tired" is much more unlikely than the animal being described with the same adjective. With regular attention, the encoding of the word "it" would only use the leftward context ("The animal didn't cross the street because") which does not help determine whether "it" refers to "The animal" or "the street". However, by looking forward and observing either "tired" or "wide", this additional context allows the attention mechanism to connect "it" with the correct entity.

Instead of using hidden states, self-attention implements a system reminiscent of the programming data structure of the dictionary or the hash table in six steps:

\begin{enumerate}
    \item Three vectors, Key, Value and Query, are calculated for each word $x_{1}, ..., x_{T}$. This is done by multiplying their respective encodings with three weight matrices $W_{K}$ (for Key) , $W_{V}$ (for Value) and $W_{Q}$ (for Query). These weight matrices are learnt during training.
    \item Each word needs a series of attention scores to determine how it relates to every other word (and itself) in the input. For the $i$-th word, this is calculated by taking the dot product of query vector individually with each key vector: $score(i, j) = q_{i} \cdot k_{j}$, calculated for each $j \in [1, ..., T]$
    \item The scores are divided by a factor of $\sqrt{d_{k}}$ where $d_{k}$ is the dimension of the key vectors. This was empirically found to produce more stable gradients \cite{attention}.
    \item Similar to the earlier form of attention, the Softmax function turns these attention scores into probabilities while also amplifying high and reducing low values.
    \item Multiplies the value vector of each word with its corresponding softmaxed score. This can intuitively be interpreted as a form of feature selection where the value vectors of important words remain rather intact while those of unimportant words are drowned out by being multiplied with scores closer to $0.001$. This produces a set of weighted value vectors $v^{(i)}_{1}, ..., v^{(i)}_{T}$ for each word $i$.
    \item Finally, the sum of weighted value vectors are calculated for each word $i$, producing $z^{(i)} = \sum v^{(i)}_{1}, ..., v^{(i)}_{T}$. This vector is the output of the self-attention layer. \cite{jalammar}
\end{enumerate}

\section{Transformer}\label{section_transformer}
Transformers are a recent deep learning method that cuts RNNs out of the architecture and relies solely on self-attention mechanisms to encode and decode, thus eliminating all of the problems inherent in RNNs. Up until this moment, RNN-based architectures like the LSTM was widely regarded as the state of the art, but with better performance, generalization and faster training, the Transformer architecture quickly dethroned RNNs and its variations in terms of training time and achieved a new state-of-the-art in English-to-German machine translation \cite{attention}.

\begin{figure}[!ht]
  \begin{center}
    \includegraphics[width=0.8\textwidth]{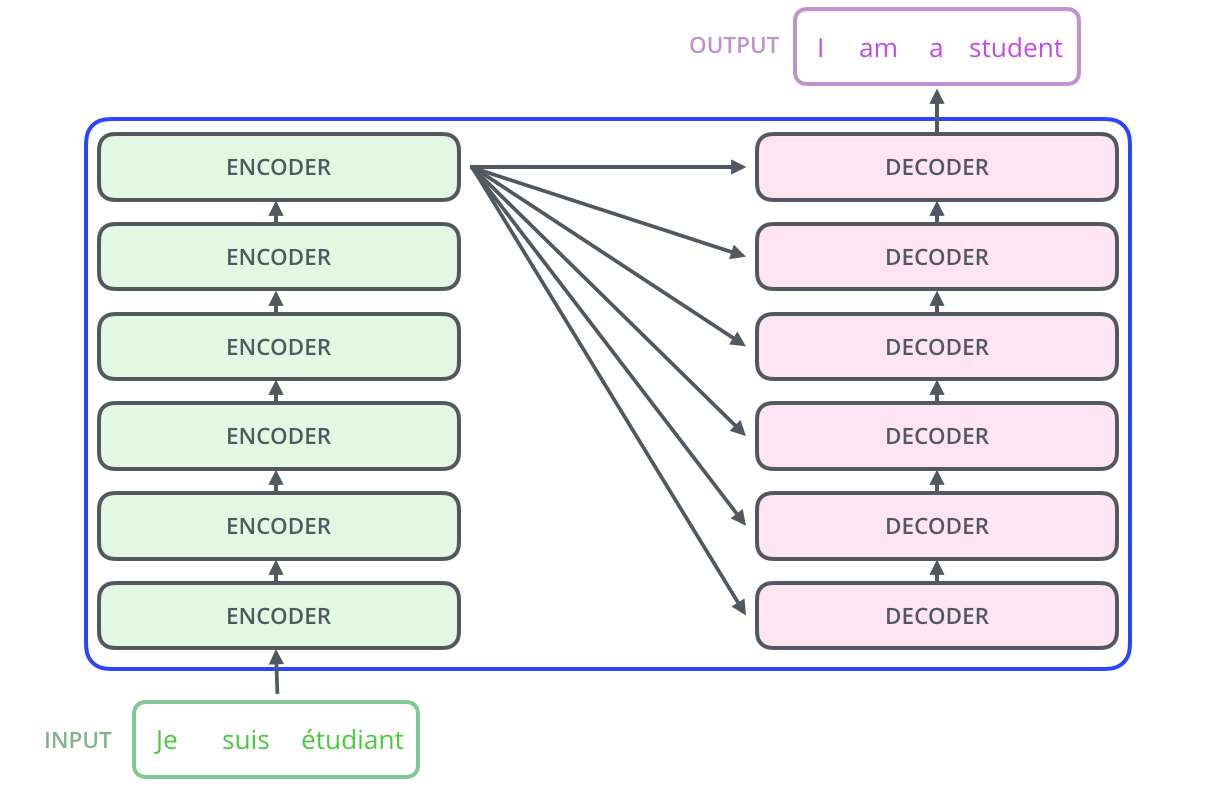}
  \end{center}
  \caption{The original Transformer architecture with six encoder blocks and six decoder blocks. From \cite{jalammar}}
  \label{fig:enc_dec_stack}
\end{figure}

The Transformer architecture, visualized in figure \ref{fig:enc_dec_stack}, consists of a number of encoder blocks stacked on top of each other and then a number (not necessarily the same as encoders) of decoder blocks also stacked on top of each other.

\subsubsection*{Embeddings}
\begin{figure}[!ht]
  \begin{center}
    \includegraphics[width=0.8\textwidth]{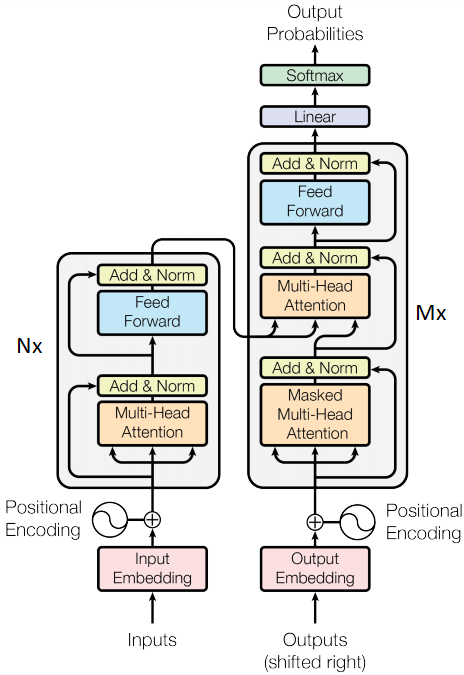}
  \end{center}
  \caption{In depth architecture of one encoder and one decoder. Adapted from \cite{attention}}
  \label{fig:transformer_low}
\end{figure}

The input is preprocessed before it is ready for the first encoder block, as seen in figure \ref{fig:transformer_low} "Input Embedding" and "Positional Encoding". First, the input sequence is embedded for reasons discussed in section \ref{sec:word_embeddings}. Then, since all words are processed simultaneously, the position of each word has to be encoded to preserve the contextual implications of word order \cite{attention}. In RNN-based architectures, this came "for free" since words were processed from left to right, on at a time, i.e. they were sent into the neural network in order.

\subsubsection*{Encoder block}
The encoder blocks are structurally identical except for their weights. As seen in figure \ref{fig:transformer_low}, the encoder consists of two sub-layers: multi-head self-attention and feed forward.

\begin{figure}[!ht]
  \begin{center}
    \includegraphics[width=0.8\textwidth]{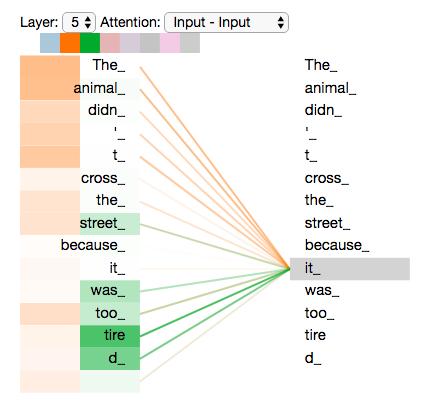}
  \end{center}
  \caption{Visualization of two attention heads (green and orange respectively) in the fifth encoder block. Darker color means more attention. From \cite{jalammar}}
  \label{fig:orange_green_attention}
\end{figure}

Multi-head attention makes it possible for the model to "divide and conquer" by specializing in different things using different weights. Each head can focus on different aspects of the problem. For example, one head could learn to couple adjectives with a subject while another could help conjugate verbs by drawing attention to the relevant context. In figure \ref{fig:orange_green_attention} we can observe that the orange attention head draws attention to mostly "The animal" and somewhat to "didn't" whereas the green focuses on "street" and "tired" among others. This means that the final encoding for the word "it" will contain its relationships to the adjective "tired", that it can refer to an animal or a street. This offers a much more nuanced view of the word "it" than could be obtained with a single attention-head. The original Transformer model uses eight (8) attention-heads, which can be run in parallel. The output of these attention-heads are then concatenated to form a singular attention vector. The attention vector is then passed into a fully-connected feed-forward layer performing a scoring calculation, applied identically and individually to each position \cite{attention}.

\subsubsection*{Decoder block}
The decoder blocks contain three sub-layers: multi-head self-attention, encoder-decoder attention and a feed-forward layer. The only structural difference from the encoder block is the "encoder-decoder attention". This layer helps the decoder focus on the relevant parts of the input sequence (received from the encoder), i.e. apply attention \cite{jalammar}.


\section{Bidirectional Encoder Representations from Transformers (BERT)}
Based on the Transformer, the 
\acrfull{BERT}
was introduced in 2018. Now only the encoder network remains and the idea is simple. The encoder is \textit{pre-trained} to handle the semantics of a language and translates the input text sequence to embeddings, just like before. What Devlin et al. (2019) realized was that this embedding can be useful for many different things. In a manner of transfer learning, the model that understands language is expensive and resource-intense to train, but its output embeddings could be used in many different tasks. The embedding can then be decoded by a model that is \textit{fine-tuned} to solve a number of NLP tasks. For example, is the sentence carrying a positive or negative message? What language is the sentence in? Which parts of the sentence constitute the main clause and which are the subordinate clause? All of these tasks depend on the same understanding of the language. This means that once an encoder is pre-trained, it can then be shared for anyone to fine-tune for their specific task, saving time and money and does not require everyone to have access to hundreds of thousands of books to build semantic understanding upon. While multilingual models exist, the best performance is most likely obtained by training a BERT model for each individual language \cite{hungarian}. For the purpose of question mark prediction, a sub-task of punctuation restoration, a BERT model can improve the F1 score from 30\% to 90\% compared to an LSTM \cite{lstm_vs_bert}.

BERT uses WordPiece tokenization
. In an effort to improve handling of rare words, the words are split into common sub-word units called wordpieces during the tokenization process \cite{wordpiece}. This can be seen in table \ref{preprocessing_examples} where the entity name Mörlunda has been spliced into Mör and \#\#lunda (an unmarked root and subsequent parts indicated with "\#\#"). With this approach, similar entity names such as Frölunda will help build semantic understanding for Mörlunda. The Swedish language is rich in compound words. The word "sjukhus" (eng. hospital) consists of "sjuk" (eng. sick) and "hus" (eng. house). If one knows the definitions of sick and house, it is easier to infer what a sickhouse implies. Thus, the data for each word is increased, since the sentence "hen var sjuk med feber" (eng. they were ill with fever) will add to the output embeddings' understanding of \textit{sjukhus} in addition to \textit{sjuk}. Additionally, it means that no other method of handling unknown words, i.e. words not observed during pre-training, is necessary since they are interpreted through sub-words \cite{wordpiece}.

BERT-based models combine three main selling points: context embeddings, transfer learning through pre-training, and having bidirectional encodings.

\subsection{Context embeddings}
Output embeddings generated by BERT are context-dependent.
In contrast to Word2vec embeddings, the same token get different embeddings depending on context (i.e. surrounding words). This allows the embeddings to capture context-based differences for words like Nice (city) and nice (good) \cite{context_dependence}.

\subsection{Pre-training}
Training a model to obtain semantic knowledge is costly. BERT was pre-trained on a combined 3.3 billion words \cite{devlin2019bert}. Not only is this a large amount of data to have access to and store, but also to train on. Training a large network is a substantial undertaking in terms of access to state-of-the-art computational power for long periods of time, which costs money in terms of electricity and rent. One solution to this is the pre-training and fine-tuning split that BERT supports.

A pre-trained model can easily be made accessible to others. As such, the cost of pre-training only has to be done once and everyone can partake in experimenting with the result for a variety of different downstream tasks. This is done through "fine-tuning" which will be covered section \ref{subsection_fine-tuning}.

The pre-training of BERT was done through two NLP tasks.

\subsubsection*{Masked language modeling}
The goal of the masked language modeling (MLM) task is to successfully predict a missing words in sentences (colloquially known as "fill in the blanks"). This task is how Devlin et al. \cite{devlin2019bert} made BERT a bidirectional encoding architecture, unlike the Transformer covered in section \ref{section_transformer}.







Devlin et al. \cite{devlin2019bert} 
randomly masked 15\% of all tokens in sentences and let the model try to predict only the masked words. The masking is primarily done by replacing the randomly selected tokens with a token "[MASK]". However, since this token will not be present in any fine-tuning project, only in pre-training, it will create a mismatch between these two processes of BERT. To lessen the effect of the presence of the mask token, masked tokens are not always replaced with "[MASK]", they can sometimes be replaced with a random token or remain unchanged instead. After testing out different shares of these three masking methods, visualized in figure , they found that 80\% [MASK], 10\% random token and 10\% remaining unchanged minimized the problematic effect.



\subsubsection*{Next sentence prediction}
The second task is based on understanding relationships between a pair of sentences, inspired by downstream tasks such as question answering (QA). Devlin et al. \cite{devlin2019bert} set up a binary prediction task where 
for a given pair of sentences A and B, predict whether or not B follows
upon A
. The sentence $B$ is the actual next sentence in 50\% of cases and a random sentence from the corpus in the remaining 50\% of cases.

\subsection{Fine-tuning}\label{subsection_fine-tuning}
The pre-trained model, that required 3.3 billion words and 4 days of training on 16 TPUs, can be downloaded as a Pytorch model of size 420.1 megabytes \cite{bert_huggingface}. This model can then be \textit{fine-tuned} to a number of different downstream tasks in just a few hours on a single GPU \cite{devlin2019bert}, such as
\begin{itemize}
    \item sequence classification,
    \item extractive question answering,
    \item language modeling,
    \item text generation,
    \item summarization,
    \item translation, and the topic of this report:
    \item named entity recognition (NER) \cite{huggingface_tasks}.
\end{itemize}

The fine-tuning often only involves adding a single output layer and training end-to-end for the specific downstream task \cite{devlin2019bert}.

\subsection{Bidirectional architecture}
BERT is bidirectional, meaning that it looks at words in the sentence both before and after the current word when calculating a label for it. This is made possible by handling the entire sentence at once rather than processing it word for word, left to right. As such, the model has a better understanding of context as it can take the entire sentence into consideration rather than just having the previous words as its observable universe. Devlin et al. \cite{devlin2019bert} argue that the unidirectionality of previous models have been a major downside in sentence-level tasks. This approach also means that BERT is parallelizable, unlike RNN and LSTM, since all words can be evaluated by attention heads in parallel. This is a substantial advantage in terms of computation time, since modern GPUs and TPUs run on thousands of cores that can execute tasks in parallel.

\subsection{Multilingual BERT}
Devlin et al. \cite{devlin2019bert} developed a multilingual BERT (mBERT) in conjunction with the regular, English BERT discussed previously. The mBERT has exactly the same architecture, but was pre-trained on a different, multilingual data set with 104 different languages represented. The model has performed "surprisingly good" (Wu and Dredze, 2020) on various NLP tasks, especially cross-lingual tasks such as translating from one language to another.

However, the multilingual approach has lackluster performance for smaller languages. Finnish researchers Virtanen et al. \cite{FinBERT} argue that there is a significant difference between high-resource and low-resource languages in how much of the pre-training corpus they constitute. Presented in table \ref{mbert_99_langs} are a small selection of the languages used for pre-training mBERT. Virtanen et al. maintain that the Finnish language was not represented enough in the pre-training of mBERT and as such the performance suffers. Romanian authors Dumitrescu et al. \cite{romanian_bert}, who have a similar resource base for their language as the Finnish, agree. Thusly they have created FinBERT \cite{FinBERT} and Romanian BERT \cite{romanian_bert} respectively and obtained multiple state-of-the-art results for their respective languages. Additionally, the next-most represented languages, German, Spanish, French and Russian, all have their respective monolingual BERT models (gottBERT \cite{gottBERT}, es-BERT \cite{es-BERT}, camemBERT \cite{camemBERT} and RuBERT \cite{RuBERT}) outperforming mBERT and sometimes attaining state-of-the-art results.

Notably, on monolingual tasks, mBERT performs worse than the English BERT, despite being the most high-resource language in the world \cite{mbert_equality}. Taking all this into account, it is not difficult to see the motivation behind making a Swedish BERT. Languages both with less and more resources have all led to successful monolingual BERT models surpassing the multilingual mBERT, despite having smaller data sets than 


\begin{table*}[!ht]
\begin{center}
\resizebox{1\linewidth}{!}{
\begin{tabular}[b]{cccc}
\toprule
log\textsubscript{2} of size in MB & Languages & \# Languages & Size Range (GB) \\
\midrule
3 & Ido (Esperanto), Sicillian & 4 & [0.006, 0.011] \\
4 & Sundanese & 6 & [0.011, 0.022] \\
5 & Irish, Icelandic & 19 & [0.022, 0.044] \\
6 & Afrikaans, Welsh & 12 & [0.044, 0.088] \\
7 & Azerbaijani, Bengali, Basque, Hindi & 15 & [0.088, 0.177] \\
8 & Bulgarian, Danish, Greek, Thai & 15 & [0.177, 0.354] \\
9 & Finnish, Norweigan & 10 & [0.354, 0.707] \\
10 & Hungarian, Dutch, Swedish, Ukrainian & 7 & [0.707, 1.414] \\
11 & Italian, Japanese, Polish, Chinese & 6 & [1.414, 2.828] \\
12 & German, Spanish, French, Russian & 4 & [2.828, 5.657] \\
14 & English & 1 & [11.314, 22.627] \\
\bottomrule
\end{tabular}
}
\caption{Some of the languages used in mBERT and its pretraining corpus size. Full table available in \cite{mbert_equality}. (Derived from \cite{mbert_equality})}
\label{mbert_99_langs}
\end{center}
\end{table*}

\subsection{KB-BERT}
Developed by the National Library of Sweden (sv. \textbf{K}ungliga \textbf{B}iblioteket), \acrshort{KBBERT} is the current state-of-the-art model for a variety of Swedish NLP tasks, notably NER tasks. The model was pre-trained in the same way as the original BERT \cite{kb-bert}.

Despite the implications of the table \ref{mbert_99_langs} putting Swedish at roughly 1/16th the size of English for the pre-training of mBERT, KB managed to amass close to 3.5 billion words \cite{kb-bert}, which is not only more than similarly "resourced" languages, but even more than Devlin et al. \cite{devlin2019bert} used when pre-training their monolingual English BERT. The data set consists of digitized newspapers, official reports from the Swedish government, legal e-deposits, social media (comments from internet forums), and the entirety of Swedish Wikipedia (detailed in table \ref{kb_corpora}). This substantial data set was made possible by the rich archives of the National Library coupled with a significant amount of preprocessing done by their data science laboratory \cite{kb-bert}.

\begin{table}[!ht]
\label{kb_corpora}
\begin{center}
\begin{tabularx}
      {\columnwidth}{X r r r}
      \textbf{Corpus}          & \textbf{Words}  & \textbf{Sentences} & \textbf{Size} \\ \hline
      Digitized newspapers     & 2 997M & 226M      & 16 783MB  \\ 
      Government texts         & 117M   & 7M        & 834MB     \\
      Legal e-deposits   & 62M    & 3.5M      & 400MB     \\
      Social media      & 31M    & 2.2M      & 163MB     \\
      Swedish Wikipedia     & 29M    & 2.1M      & 161MB     \\ \hline
      Total           & 3 497M & 260M      & 18 341MB  \\
\end{tabularx}
\caption{Size of cleaned corpora used when pre-training KB-BERT. (Derived from \cite{kb-bert})}
\end{center}
\end{table}

Only a cased version of KB-BERT exists, which is sub-optimal for the purpose of this report since the data will be uncased and a mismatch is grounds for performance drops \cite{hungarian}.

\section{Named entity recognition (NER)}
\Acrlong{NER} is a classification task where tokens are individually classified as belonging to different "named entities" such as an organisation, location or personal name. The nine classes often used in NER tasks are:

\begin{itemize}
    \item O, Outside of a named entity

\item B-MIS, Beginning of a miscellaneous entity right after another miscellaneous entity

\item I-MIS, Miscellaneous entity

\item B-PER, Beginning of a person’s name right after another person’s name

\item I-PER, Person’s name

\item B-ORG, Beginning of an organisation right after another organisation

\item I-ORG, Organisation

\item B-LOC, Beginning of a location right after another location

\item I-LOC, Location \cite{huggingface_tasks}
\end{itemize}

These classes support multiple words belonging to the same entity, such as "New York City" being one location. An example of NER tagging can be seen in table \ref{NER_example}.

\begin{table}[]
\centering
\begin{tabular}{|c|c|c|}
\hline
Token     & Class & Class (semantic) \\ \hline
Hu        & I-ORG & Organisation     \\ \hline
\#\#gging & I-ORG & Organisation     \\ \hline
Face      & I-ORG & Organisation     \\ \hline
Inc       & I-ORG & Organisation     \\ \hline
.         & O     & None             \\ \hline
is        & O     & None             \\ \hline
a         & O     & None             \\ \hline
company   & O     & None             \\ \hline
based     & O     & None             \\ \hline
in        & O     & None             \\ \hline
New       & I-LOC & Location         \\ \hline
York      & I-LOC & Location         \\ \hline
City      & I-LOC & Location         \\ \hline
.         & O     & None             \\ \hline
\end{tabular}
\caption{Example of NER BERT predictions. From \cite{huggingface_tasks}}
\label{NER_example}
\end{table}


\subsection*{Punctuation restoration}\label{sec:punctuation_restoration}
Punctuation restoration is the task of adding punctuation (such as periods, commas, question marks etc.) to a text. The "restoration" part of the name may imply that it is only used in cases where a piece of text has somehow lost its punctuation, such as an imperfect optical character reader (OCR) or speech-to-text where punctuation is implicit. However, nothing is stopping us from using exactly the same model for \textit{introducing} punctuation into a computer-generated text, even if there is no "definitive" answer.

Capitalization is strongly correlated with punctuation. For example, the first letter in a sentence is always capitalized, meaning that period and capital letters often appear together. For this reason, Nagy et al. \cite{hungarian} chose to turn the entire data set into lowercase during the preprocessing step in order to eliminate this bias. This approach fits that of this report, where the real-world problem to solve is unpunctuated (alternatively "uncased").

Punctuation restoration can be formulated as a NER task by assigning each word a label of the punctuation immediately following it. There are often only four classes: PERIOD, COMMA, QUESTION, and OTHER/EMPTY where the latter is used for marking no punctuation \cite{adversarial, alam, hungarian}. For example, the sentence "susan, where is the national library?" would get the labels "COMMA EMPTY EMPTY EMPTY EMPTY QUESTION". There are other forms of punctuation, such as semicolons, dashes, exclamation points etc. but these are too seldomly seen in text to meaningfully train on. As such, it is often decided to stick to these four classes and merge other forms of punctuation into the three overarching classes. For example, exclamation points are used according the same grammatical rules as period, albeit with a certain dramatic flair. It is therefore possible to convert all exclamation points into periods during preprocessing to get more meaningful data \cite{hungarian}. This concept will be thoroughly explored in section \ref{sec:preprocessing}.

Even a grammatically correct source text can be ambiguous. After all, part of why punctuated text is easier to interpret \cite{tundik_readability} is because it clears up ambiguity. For example, the unpunctuated sentence "solen skiner idag är det varmt" has multiple different valid interpretations:
\begin{itemize}
    \item solen skiner. idag är det varmt. (eng. the sun shines. it is hot today)
    \item solen skiner, idag är det varmt. (eng. the sun is shining, today it is hot.)
    \item solen skiner idag. är det varmt? (eng. the sun is shining today. is it hot?)
\end{itemize}

One worry is harsh penalties during testing as different interpretations can essentially result in "double-counted" errors. For example, if the correct sentence is "... skiner. idag ..." and the prediction is "... skiner idag. ...", both the words "skiner" and "idag" will have been classified wrong, adding one false positive and one false negative to the PERIOD class. However, recent BERT-based models have had substantial success in achieving state-of-the-art performance despite disregarding this ambiguity problem altogether. One can speculate that the ambiguity is (1) relatively rare and that most text does not have more than one or two valid interpretations and (2) that the model learns to mimic the punctuation style of the training set and that end-to-end approach is sufficient to avoid the aformentioned harsh penalties during testing.

\section{F1-score}
\begin{figure}[!ht]
  \begin{center}
    \includegraphics[width=0.8\textwidth]{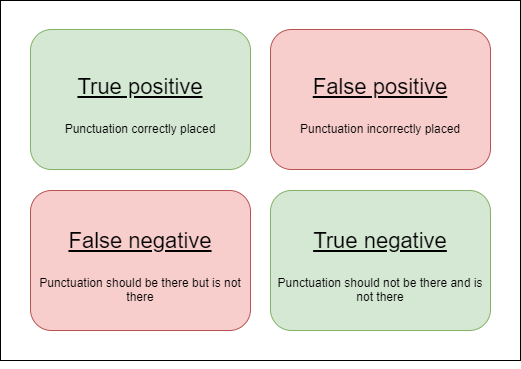}
  \end{center}
  \caption{2D schema of positives and negatives}
  \label{2dposneg}
\end{figure}

For each class EMPTY, PERIOD, COMMA and QUESTION, model estimates can be considered a part of one of the following four groups: true positive, false positive, true negative and false negative, illustrated in figure \ref{2dposneg}. The same prediction is put into different categories depending on point of view. The definitions below use the question mark class QUESTION as an example.

True positives (TP) are tokens where prestoBERT has predicted QUESTION and the label from the original text source material is also QUESTION. Example: The question mark in "är allt bra?" is in the original sentence and in the predicted sentence. This means that the prediction for "bra" is a true positive.

False positives (FP) are tokens where prestoBERT has made an erroneous prediction QUESTION where the correct label is not QUESTION. Example: "hallå jag? är hemma" where the correct label for "jag" should be EMPTY.

True negatives (TN) are tokens which prestoBERT correctly classified as not belonging to QUESTION. Example: "hallå jag. är hemma". In the example, none of the words belong to the QUESTION class. Even though "jag" still has the wrong class, from the perspective QUESTION, it is a success.

False negatives (FN) are tokens where prestoBERT has made an erroneous prediction that should have been QUESTION. Example: "var det bra så" where "så" has the predicted label EMPTY but the true label is QUESTION.

Using these for categories, some measurements for ML models include accuracy, precision, recall and F1-score.

A commonly used measurement for machine learning model performance is accuracy. It measures the proportion of true positives and true negatives out of the total population, see \ref{acc_eq}. However, even though the equation contains every group (TP, FP, TN and FN), accuracy lacks nuance. For example, if the data set consists of 93\% EMPTY labels, having a simple model that always predicts EMPTY will yield 93\% accuracy. This highlights one of two problems with accuracy: class imbalance. In the data set, the distribution of classes, i.e. the number of each punctuation mark and EMPTY, is heavily skewed towards the EMPTY class. 

In the context of punctuation restoration, the purpose can range from increasing readability to reaching an ultimate level of grammatical correctness. In the former case, we could consider the sentence "det blev allt eftersom bättre". There are two commas missing, but it is quite readable as is. Depending on how we measure success, we might end up with anything from "det, blev, allt, eftersom, bättre" and "det blev, allt eftersom bättre", neither which are entirely correct, but for different reasons.

Precision is a measurement of many of our class predictions are correct. In this context: "of all the commas we predicted, how many are correct?". This measurement excludes FN, which makes it favorable to "not guess" if the model is unsure. In some contexts, this is especially important. When ranking search results, it might be more important to get a good 10 results rather than 20 good ones and 5 bad ones. Precision essentially measures the quality of the positive cases and is defined in equation \ref{precision_eq}. This measurement could likely result in the prediction "det blev, allt eftersom bättre" where only one of the commas are predicted, because there is a penalty for predicting a comma where it should not be, but no penalty for missing a comma that should be there. To get a perfect precision, the model only needs to predict one comma correct and not venture another guess from there on out.

Recall is in some ways the mirror image of precision. By swapping out FP to FN in the formula, it does not pay any attention to wrongful predictions of comma but rather "how many commas were left out". To get a perfect recall, the model only has to maximize TP commas. A shortcut to that is to predict comma for every label, which corresponds to the case "det, blev, allt, eftersom, bättre". This is not desirable either. In the context of search results, it would equate to returning all relevant results, buried in a mountain of irrelevant results. However, recall could be seen as more important than precision in the case of cancer diagnosis where the penalty for not discovering cancer in a patient is often death, but finding FP is only the cost of a second test.

The F-score combines both precision and recall into a single score between 0 and 1, with 1 being a perfect performance. This gives a more nuanced measurement for cases where accuracy is inadequate due to aforementioned reasons. There are many ways to weigh precision and recall differently, but a common choice among similar papers is to use the harmonic mean, named the F1-score, which weighs both factors equally.

For comparison between different models, the macro F1-score can be appropriate as a summary of overall performance \cite{hungarian}. The macro F1-score, as defined by Scikit-learn, is the unweighted average of F1-scores \cite{macro_f1}. 

\begin{equation}\label{acc_eq}
    accuracy = \dfrac{TP+TN}{TP+FP+TN+FN}
\end{equation}
\begin{equation}\label{precision_eq}
    precision = \dfrac{TP}{TP+FP}
\end{equation}
\begin{equation}\label{recall_eq}
    recall = \dfrac{TP}{TP+FN}
\end{equation}
\begin{equation}\label{f1_class_eq}
    F1(class) = 2\cdot\dfrac{precision(class)\cdot recall(class)}{precision(class)+recall(class)}
\end{equation}
\begin{equation}\label{f1_eq}
    F1\textnormal{-}score = \dfrac{\sum_{class}{F1(class)}}{N}
\end{equation}
\begin{center}
    where $class \in \{EMPTY, PERIOD, COMMA, QUESTION\}$\newline
\end{center}

\cleardoublepage
\chapter{Method}
\section{Data set}
A data set consisting of government texts such as newsletters from different counties and municipalities in Sweden was used in this project. These texts are rather homogeneous since they are written in official and formal Swedish. As a consequence, the entire data set usable from a grammatical standpoint since it has been proofread. In contrast, the creators of KB-BERT describe the usage of punctuation in text originating from social media as being seemingly optional \cite{kb-bert}. This makes such a data set nonindicative of how punctuation is used in Swedish, especially in this context where punctuation is the focus. However, this homogeneity has its downsides as question marks are heavily underrepresented. The amount of Swedish data used for fine-tuning can be seen in table \ref{dataset_breakdown}.

Compared to the Hungarian study, this data set is significantly smaller (see: table \ref{szeged_vs_swe} and figure \ref{bar_graph_dataset}). Overall, it is only about 30\% as large as the Szeged treebank data set \cite{hungarian} and consists of a higher frequency of the EMPTY class, thus offering an even more skewed data set with fewer examples of the rare classes.


\begin{table}[]
\centering
\begin{tabular}{|l|l|l|l|}
\hline
Data set        & Full set & Training set & Test set \\ \hline
\# of documents & 301      & $\sim$241    & $\sim$60 \\ \hline
\# words        & 371,973  & 298,872      & 73,101   \\ \hline
\# PERIOD       & 19,822   & 15,928       & 3894     \\ \hline
\# COMMA        & 10,911   & 8909         & 2002     \\ \hline
\# QUESTION     & 44       & 37           & 7        \\ \hline
\# EMPTY        & 341,196  & 273,998      & 67,198   \\ \hline
\end{tabular}
\label{dataset_breakdown}
\caption{Breakdown of the Swedish data set used in the fine-tuning of prestoBERT.}
\end{table}

\subsection{Preprocessing}\label{sec:preprocessing}
For the purposes of the overarching project at KTH, punctuation restoration is to be done on lower-cased text. Thus, we simply exchange each upper-case letter for its lower-case counterpart.

As argued by Nagy et al., some types of punctuation are too seldomly used to get enough samples to reliably learn their patterns. These include semicolons, exclamation points, dashes etc. Many of these are also interchangeable with the more common forms of punctuation. For example, exclamation points are only a niche form of period. They both end sentences, although exclamation points do so with a certain semantic tone. Grammatically, however, they are equivalent. As such, these rare forms of punctuation are absorbed into three large groups: periods, commas, and question marks. This means that, for example, each exclamation point in the data set has been replaced by a period.

Lastly, due to the nature of how the data was collected (likely in pdf format), some text is contains dashes as mid-word linebreaks. This is a side-effect of being fitted to a specific width. To preserve the actual words, all instances of a dash followed by a newline are removed to merge affected words back together. A full list of changes can be seen in table \ref{preprocessing}.

\subsection{Feeding into BERT}\label{sec:feeding_into_BERT}
Since the problem is posed to BERT in the form of token classification (a simpler form of named entity recognition), each token in the training and test set requires a label indicating its true class. The classes selected are PERIOD, COMMA, QUESTION and EMPTY, the latter being the class for everything else. We can not feed the actual punctuation into BERT, since then BERT will only learn to classify "." as PERIOD, which is trivial and not solving the problem at hand. Instead, each word preceding a form punctuation is assigned the corresponding class, see table \ref{preprocessing_examples} for examples.

The result are two large vectors consisting of words and their corresponding tags respectively. However, sending each word into BERT on its own would not allow BERT to utilize its attention mechanism. Rather, the input to BERT needs to be in the form of entire sentences. Feeding in one sentence at a time has its own problem: it is trivial to predict the period at the end. To combat this problem, sentences are fed into BERT between three to seven sentences (henceforth referred to as "compound sentences") at a time. The number is chosen randomly. This still introduces bias as every 3rd to 7th period prediction will be trivial. This will be taken into consideration in the results chapter. Another solution would be to truncate the text into batches of 512 tokens, but this, too, has issues, which will be discussed in section \ref{sec:disc:prestoBERT}.

Similar experiments shuffled the order of sentences to make punctuation more evenly distributed \cite{hungarian}, e.g. a transcript of a questionnaire might contain many more question marks than a news article. This has the downside of introducing a form of bias in the neural network. Shuffling the sentences means that the semantic context changes rapidly between pairs of sentences, since they are drawn from very different sources. This makes it possible for the neural network to lean too heavily on discerning a period based on how the context changes. The effect of this in the training and test sets is significantly bigger than in a real-world application \cite{hungarian}. As previously mentioned, the Swedish data set is significantly more homogeneous, which makes the shuffling unnecessary and avoids introducing this form of bias.

The data is split 80/20 into the training and test set respectively. This number is a common but ultimately arbitrary choice in machine learning practice. No source with empirical evidence could be located.

\subsubsection{Tokenization}
The data is tokenized using the built-in tokenizer of KB-BERT. Since each token encoding needs its own index, some words (e.g. particularly long or rare words like compound words and enitity names) are split up into multiple tokens, exemplified by Mörlunda becoming "mör" and "\#\#lunda" in table \ref{preprocessing_examples}. Tokens prefixed with "\#\#" are spliced from the root. This is important to keep track of for correctly tagging tokens.

Additionally, two special tokens are appended to the beginning and end of each compound sentence. These have more intricate functionalities in more advanced tasks and thus might seem superfluous in this context. The tokens are the classification token (CLS) and separation token (SEP).

CLS is a token reserved for an aggregate, compound sentence-wide class estimate. For example, if the task was to determine if the input was in Swedish or not, the CLS token be transformed into a class prediction in the output (in that case 0 or 1) \cite{devlin2019bert}. In this context, prediction is carried out on a token-by-token basis with no overarching classification, rendering the CLS token irrelevant.

The SEP token is used to separate different parts of the input. In more advanced problems, for example involving questions and answers, SEP is used to separate these distinct parts of the input. In this context, it signals the end of the input compound sentence and where potential padding begins. This is a by-product of every sentence fed into BERT having to be of the same length. Since the lengths of compound sentences vary greatly, most are padded with dummy tokens, i.e. meaningless tokens ignored when evaluating the input. The SEP token indicates where the padding starts so that it can be ignored. Otherwise, the padding would be treated as the same word repeated until for the remainder up until maximum length. The maximum length was initially set to the longest compound sentence after encoding, but this proved unstable. Instead, the maximum was set to 512 encoded tokens, as recommended by \cite{devlin2019bert}. For this problem, truncating sentences exceeding 512 tokens creates issues: what would the tag for "mör" be if "\#\#lunda" is truncated? Instead, compound sentences in need of truncation are removed from the feed. In practice, despite being dependent on randomness in which sentences are concatenated, there tends to be only one compound sentence in the data set exceeding the limit. The effect of this harsh truncation is thus negligible.

\begin{table}[]
\centering
\begin{adjustbox}{angle=90}
\label{preprocessing_examples}
\begin{tabular}{l|llllll}
\hline
Original     & Mörlunda &      & stationer, & samt & de & icke! \\ \hline
Preprocessed & mörlunda &      & stationer, & samt & de & icke.  \\
Tokenized    & mör & \#\#lunda & stationer  & samt & de    & icke   \\
Output       & EMPTY    & -    & COMMA      & EMPTY & EMPTY & PERIOD \\ \hline
Original     & aldrig & - & oavsett & vad & - & igen.             \\ \hline
Preprocessed & aldrig, & oavsett & vad, & igen.        &        \\
Tokenized    & aldrig                            & oavsett   & vad       & igen   &       &        \\
Output       & COMMA                             & EMPTY     & COMMA     & PERIOD &       &        \\ \hline
\end{tabular}
\end{adjustbox}
\caption{Examples of preprocessing and feeding sentences into BERT}
\end{table}

\begin{table}[]
\centering
\begin{tabular}{|c|c|}
\hline
Original                 & Preprocessed \\ \hline
A ... Ö                  & a ... ö      \\ \hline
\#                       &              \\ \hline
-                        & ,            \\ \hline
-\textbackslash{}newline &              \\ \hline
;                        & :            \\ \hline
!                        & .            \\ \hline
"                        & ,            \\ \hline
\end{tabular}
\caption{Preprocessing changes}
\label{preprocessing}
\end{table}

\begin{table}[]
\centering
\label{szeged_vs_swe}
\begin{tabular}{|l|l|l|l|}
\hline
Punctuation & \# in Hungarian & \# in Swedish & percent (SE \textdiv HU) \\ \hline
PERIOD      & 93,756           & 19,822         & 21.1                                \\ \hline
COMMA       & 138,693          & 10,911         & 7.9                                 \\ \hline
QUESTION    & 2,081            & 44            & 2.1                                 \\ \hline
EMPTY       & 1,023,183         & 341,196        & 33.3                                \\ \hline
Total       & 1,257,713         & 371,973        & 29.6                                \\ \hline
\end{tabular}
\caption{Comparing the Hungarian and Swedish data sets}
\end{table}

\section{BERT}
KB-BERT is a pre-trained BERT model, but it is incomplete for solving any language problem.

\subsection{Custom classification layer}
\begin{figure}[!ht]
  \begin{center}
    \includegraphics[width=0.8\textwidth]{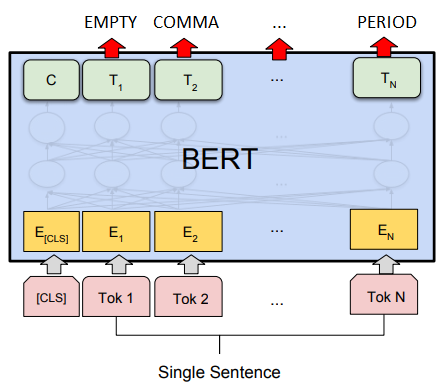}
  \end{center}
  \caption{Architecture for Single Sentence Tagging Task to be fine-tuned (based on \cite{devlin2019bert})}
  \label{single_sentence_tagging_task_architecture}
\end{figure}

For any task, be it classification, text generation or otherwise, the underlying BERT model needs to be appended with a custom output layer to pool the penultimate layer output into the number of classes used for the particular problem \cite{devlin2019bert}. In this context, there are four classes (PERIOD, COMMA, QUESTION and EMPTY). The structure of this layer was heavily inspired by Huggingface's BertForTokenClassification and token classification script run\_ner.py \cite{BertForTokenClassification, run_ner}. A graphical overview can be seen in figure \ref{single_sentence_tagging_task_architecture}.

The predictions can not be made on a word-by-word basis, since some words have been spliced into two or more WordPiece-tokens, e.g. "Mörlunda" becoming "Mör" and "\#\#lunda" in table \ref{preprocessing_examples}. To predict a label for the word "Mörlunda", there are a number options. Predict one label per spliced piece (root token and subsequent sub-word tokens) and pool their results in some way, predict only on the first token or only on the last token. Nagy et al. \cite{hungarian} chose to predict only on the last token. In their preprocessing, "tyrannosaurus" tokens "ty" and "\#\#rano" get no label while "\#\#saurus" gets an EMPTY label. In contrast, the \acrshort{NER} fine-tuning guide from HuggingFace \cite{huggingface_tokner} only labels the root token. This implementation has been done according to the HuggingFace method, but there is reason to investigate the performance of the other methods, too, which will be discussed in section \ref{sec:disc:prestoBERT}.

During preprocessing when each word is assigned its true tag. Then, during the tokenization process when each word is tokenized into WordPiece tokens, each sub-word token is masked with the label -100 while the root token receives the true tag. For example, the numeric label for QUESTION is 3, so the input "Mörlunda?" would result in the set of tokens ['Mör', '\#\#lunda'] and a corresponding tag set [3, -100]. The model still calculates predictions for each token, but the output tokens with value -100 are not included in the loss calculations, effectively discarding them. As a consequence, this model has not trained on sub-word token prediction, only root token prediction.

\subsection{Fine-tuning}
The classification layer also needs to be trained so that the underlying BERT model can learn what distinguishes the different classes. For example, if the four classes were subject, verb, adjective and preposition, the classification would look very different. This process is called fine-tuning.

The fine-tuning was done using the Huggingface trainer class on one Tesla T4 GPU provided by Google Colab through cloud computing. The loss function was cross-entropy loss. Training went on for 4 epochs and took 27 minutes and 17 seconds. 
Hyperparameters include using the Adam optimizer \cite{adam} with an initial learning rate of $10^{-5}$, a batch size of 4 per device both for training and evaluation and 0 warm-up steps.

\section{Baseline method}
Evaluating a machine learning model requires some point of reference. For example, a 60\% accuracy is low if it is in the context of binary classification, but if it is predicting the outcome of an unbiased coin flip, it is significantly better than a naive solution. A baseline method is often a simple heuristic approach to solving the problem. They serve as a point of reference to how difficult the problem is to solve. The baseline method used in this report is a human evaluation. 


\subsection{Human evaluation}\label{sec:method:human}
Literate humans have spent much time consuming written content and as such, we can assume that they serve as an adequate point of reference as to how difficult punctuation restoration is for this particular data set. Additionally, it might bring evidence to support comparison to other data sets by indicating that this data set is non-trivial.

Assuming that around 10-15 minutes of work is appropriate for unpaid test takers, a trial test of 6500 words was done by myself to gauge how long the task would take. After 13 minutes, only about 650 words had been processed. Human test sets were therefore divided into 650 words per test. These words are taken in the same order that prestoBERT received them, i.e. reading left to right without shuffling any documents. These documents are devoid of the punctuation removed in the preprocessing step. A sample of a test sent to a human test taker can be seen in appendix \ref{appendix_sample_test}.

Keeping the 650 word limit exact for all tests realistically requires some truncating, since few continuous sentences amount to exactly 650 words. As a consequence, the sentences in the beginning and end of a test will likely be missing words. This is not ideal, since it does not correspond to how prestoBERT saw the sentences (3-7 full sentences at a time). However, the alternative of splitting the data based on number of sentence (trying to make it close to 650 words) would introduce another form of bias. Some participants with longer texts would have a higher degree of influence compared to others and it is likely that the test becomes easier the more information is observable, since the participant might start to notice that there can be rubrics, itemized lists etc. in the test.

After a test document has been corrected by a human participant, the tags for each word are calculated in the same manner as the original source text. The human tags are then compared to the true tags from the source text.

The entire test set contains 375,214 words, which means that roughly 112 participants are needed to cover the entire test set. This is practically infeasible for this project. 16 human evaluations were gathered from native speakers from a variety of ages above 18, occupations and geographical location. Each person was assigned one unique test of 650 words and asked to add periods, commas and question marks as they saw fit in the text. The instructions attached to the test in included in appendix \ref{appendix_instructions}.

\chapter{Results and Analysis}
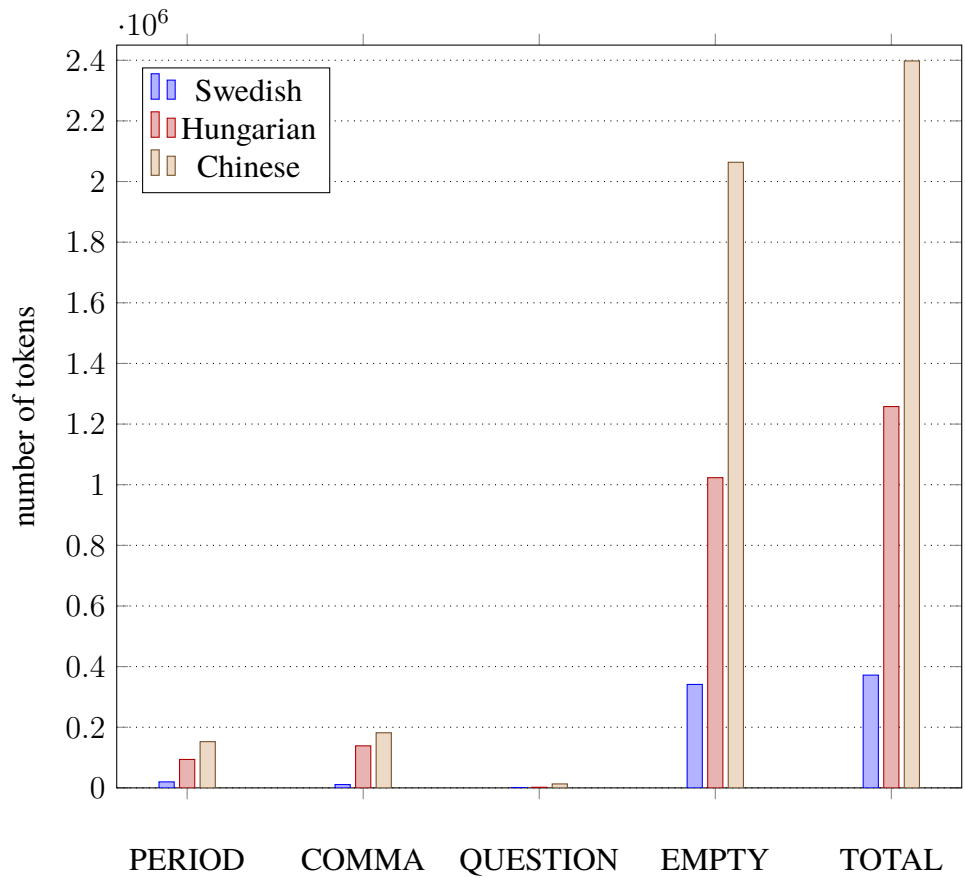
\begin{figure}
\centering
\begin{tikzpicture}
\pgfplotstableread{
class  A  B  C

PERIOD 19822 93756 152428

COMMA 10911 138693 181804

QUESTION 44 2081 13059

EMPTY 341196 1023183 2063552

TOTAL 371973 1257713 2397784

}\mytable

\begin{axis}[
    ybar,
    width=5in,
    height=4.5in,
    ymin=0,
    ymax=2450000,
    bar width=0.2cm,
    ymajorgrids=true,
    major grid style={dotted,black},
    ylabel={number of tokens},
    symbolic x coords={PERIOD, COMMA, QUESTION, EMPTY, TOTAL},
    xtick=data,
    x tick label style={yshift=-0.5cm,name=xlabel\ticknum},
    legend pos=north west
  ]

  \addplot table [x=class, y=A, draw=black, fill=blue!20] {\mytable};
  \addplot table [x=class, y=B, draw=black, fill=blue!20] {\mytable};
  \addplot table [x=class, y=C, draw=black, fill=blue!20] {\mytable};

\legend{Swedish, Hungarian, Chinese}
\end{axis}
\end{tikzpicture}
\label{bar_graph_dataset}
\caption{Comparing the data sets of punctuation restoration projects for different languages. Hungarian: \cite{hungarian}, Chinese: \cite{adversarial}.}

\end{figure}

Despite using a small data set (see comparison in figure \ref{bar_graph_dataset}), prestoBERT performed well when compared to similar non-English projects and the human evaluation. All results are presented in table \ref{table:results}. BERT-base-multilang-cased and Hubert are from Nagy et al. \cite{hungarian} and are evaluated on the Hungarian Szeged Treebank data set. BERT-BLSTM-CRF is from Yi et al. \cite{adversarial} and evaluated on a Chinese data set of manually transcribed audio files, provided by IWSLT2011. The human evaluation and prestoBERT are evaluated on the same test set, albeit only a small part of it for the human evalutation.

\newcommand{\rescolwidth}{0.05}
\begin{table*}[!ht]
\centering
\begin{adjustbox}{angle=90}
\small
    \centering
    \begin{tabular}{p{0.28\textwidth}|
            >{\centering}p{\rescolwidth\textwidth}
            >{\centering}p{\rescolwidth\textwidth}
            >{\centering}p{\rescolwidth\textwidth}
            |
            >{\centering}p{\rescolwidth\textwidth}
            >{\centering}p{\rescolwidth\textwidth}
            >{\centering}p{\rescolwidth\textwidth}
            |
            >{\centering}p{\rescolwidth\textwidth}
            >{\centering}p{\rescolwidth\textwidth}
            >{\centering}p{\rescolwidth\textwidth}
            |
            >{\centering}p{\rescolwidth\textwidth}
            >{\centering}p{\rescolwidth\textwidth}
            c
    }
        \toprule
        & \multicolumn{3}{c}{Comma} & \multicolumn{3}{c}{Period} & \multicolumn{3}{c}{Question} & \multicolumn{3}{c}{Overall}\\
        Models & P & R & $F_1$ & P & R & $F_1$ & P & R & $F_1$ & P & R & $F_1$ \\
        \midrule
        Human evalution & 59.8 & 63.3 & 61.5 & 87.2 & 80.4 & 83.6 & 100.0 & 100.0 & 100.0 & 82.3 & 81.2 & 81.7 \\
        prestoBERT (cased) & 79.2 & 64.2 & 70.9 & \textbf{90.2} & 89.3 & 89.7 & 72.4 & \textbf{79.0} & \textbf{76.0} & 80.6 & 77.5 & 78.9 \\
        BERT-base-multilang-cased \cite{hungarian} & 81.3 & 79.3 & 80.3 & 82.4 & 83.2 & 82.8 & 51.6 & 21.3 & 30.2 & 71.8 & 61.3 & 64.4 \\
        Hubert (Hungarian) \cite{hungarian} & \textbf{84.4} & \textbf{87.3} & \textbf{85.8} & 89.0 & \textbf{93.1} & \textbf{91.0} & 73.5 & 66.7 & 69.9 & \textbf{82.3} & \textbf{82.4} & \textbf{82.2} \\
        BERT-BLSTM-CRF (pre-trained, Chinese) \cite{adversarial} & 74.2 & 69.7 & 71.9 & 84.6 & 79.2 & 81.8 & \textbf{76.0} & 70.4 & 73.1 & 78.3 & 73.1 & 75.6 \\
        \bottomrule
    \end{tabular}
    \end{adjustbox}
    \caption{Precision, recall and $F_1$-score values on various data sets and languages. Best non-human results per column in bold.}
    \label{table:results}
\end{table*}

\section{prestoBERT}\label{res:prestoBERT}
The performance of prestoBERT compares favorably to other methods, as seen in table \ref{table:results}. The more advanced approach of both the Hungarian Hubert and Chinese BERT-BLSTM-CRF does not seem to have a significant positive impact. Hubert surpasses prestoBERT in almost all categories, but conversely, prestoBERT in turn exceeds the Chinese model in almost all categories, despite the Chinese model being arguably the most complex. This could be because the Chinese punctuation restoration problem is more difficult with Chinese characters than western letter-based texts, as argued in their report. It is noteworthy that both the Chinese and Hungarian projects had substantially more data, which could be the leading cause in the shortcomings of prestoBERT.

The computation time for prestoBERT's predictions was 16 seconds. For a single human to punctuate the same amount of data, it would take roughly 112 times 15 minutes, which is 28 hours. It is therefore possible to argue that prestoBERT would have real-world applications even if it performed worse than the human baseline due to the substantial time savings, taking roughly a 6300th of the time.

\begin{figure} 
    \centering
    \includegraphics[width=0.9\textwidth]{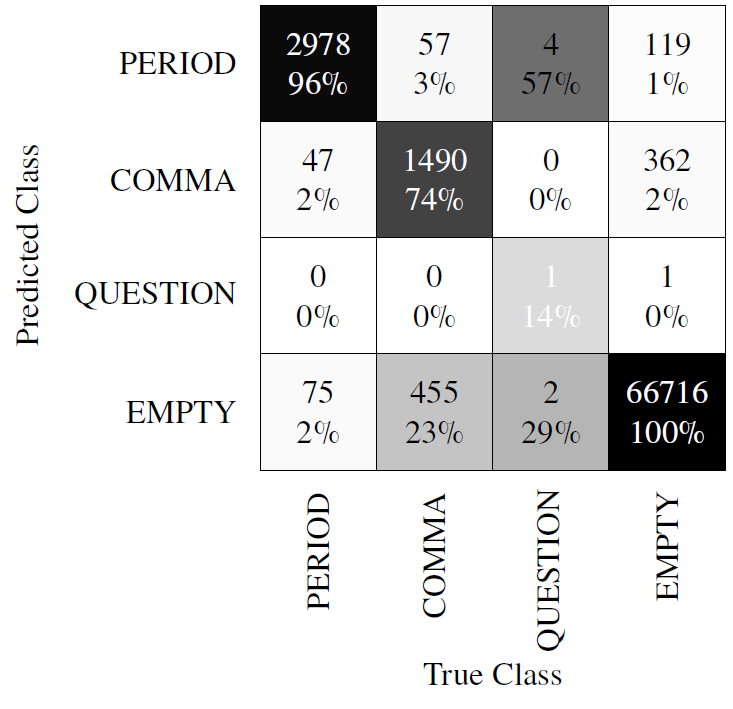}
    \label{confusion_presto}
    \caption{Confusion matrix for prestoBERT predictions (not the same data configuration of batches as the original test run in table \ref{table:results}). Free period prediction due to batch end have been disregarded in the calculation (see section \ref{sec:batch-induced_bias}}
\end{figure}

The confusion matrix is presented in figure \ref{confusion_presto}.
The performance for "QUESTION" tokens here is significantly different compared to the results presented in table \ref{table:results}. This is likely due to the low sample size of seven question marks in both the training and test sets. As a consequence, prestoBERT has likely not learned enough about question marks to correctly predict them, nor has it been sufficiently tested to provide a robust measurement on its "QUESTION" performance. As such, the result for this small class should be disregarded entirely. We can also observe that "COMMA" is frequently mistaken for "EMPTY". This might stem from commas sometimes being voluntary and there might be many grammatically sound interpretations. Section \ref{sec:old_fiction} shows an example where commas can be ignored. Looking at the column for "EMPTY", the number of false negatives attributed to "COMMA" is close to the number of "COMMA" falsely attributed to "EMPTY". This could mean that commas are specially related to "EMPTY", for example that they are often interchangeable (optional commas) or that they are swapped around (i.e. that the number of commas in a sentence is similar, but they are in the "wrong places" compared to the true labels).

The sum of false positive "EMPTY" labels is
\begin{equation}\label{eq:FN_EMPTY}
    FP_{EMPTY} = 75+455+2 = 532
\end{equation}

and the sum of false negative "EMPTY" labels is

\begin{equation}\label{eq:FP_EMPTY}
    FN_{EMPTY} = 119+362+1 = 482.
\end{equation}

The difference of only 50 cases, in the context of over 71 000 labels, indicates that prestoBERT does not err on too much punctuation nor too little.

\subsection{Batch-induced bias}\label{sec:batch-induced_bias}
As previously mentioned in section \ref{sec:feeding_into_BERT}, there is a stochastic degree of bias introduced in period prediction since compound sentences (batches) always end with a period. This makes every 3rd to 7th period trivial to predict since prestoBERT quickly learns that the token before "[SEP]" is always a "PERIOD". This affects the results in two ways: (1) the number of true positive period predictions increases by the number of compound sentences in the test set. (2) The guaranteed period helps prestoBERT 
to conclude some additional facts about the structure of the input.
For example, it is substantially unlikely that there are two periods in a row, so the 3rd last token, before "PERIOD" and "[SEP]", is unlikely to become a false positive "PERIOD". This bias is difficult to track.

To correct for the first part of the bias, the trivial periods can simply be detracted from the result, as they have been in figure \ref{confusion_presto}. The effect, however, is minimal. The number of true positive "PERIOD" before were 3772 (97\% precision) and 2978 (96\% precision) afterwards.

The second part of this bias can not be calculated and will instead be discussed in chapter \ref{ch:discussion}.

\section{Human evaluation}
Since 16 people participated in the study, only 16/112 of the data set could be covered, which is roughly 15\%. Thus, the results (seen in \ref{table:results}) should be interpreted with the caution that low sample sizes require. The human participants performed significantly worse compared to prestoBERT, except for the "QUESTION" class. This statistic should be ignored since the sample size is too small with only one question mark being presence in the human slice of the test set. However, it is still noteworthy that there were no false positive question marks. The participants were told that question marks could be present, just like periods and commas, but there were no indication given that there would be particularly few, as seen in the instructions shown in appendix \ref{appendix_instructions}. This is a substantial improvement over every other BERT model features in table \ref{table:results}. This indicates that the human participants have some semantic understanding that is harder for the model to build during training. This is probably because they have seen far more than 44 question marks in their life. In contrast, the human participants had a similarly hard time with commas as prestoBERT, as seen in the confusion matrix \ref{confusion_human}, often mistaking them for the "EMPTY" class. In all columns but "QUESTION", the humans performed worse than prestoBERT.

\begin{table}[!ht]
\centering
\label{human_eval_test}
\begin{tabular}{|l|l|l|}
\hline
Punctuation & \# in human test set & \# in full test set \\ \hline
COMMA       & 313 & 2002    \\ \hline
PERIOD      & 652 & 3894  \\ \hline
QUESTION    & 1 & 7     \\ \hline
EMPTY       & 10084 & 67198\\ \hline
Total       & 11050 & 73101 \\ \hline
\end{tabular}
\caption{True label statistics for the test set manually labeled by human participants.}
\end{table}

\begin{figure} 
    \centering
\def\myConfMat{{
{524, 11, 0, 66},
{52, 198, 0, 81},
{0, 0, 1, 0},
{76, 104, 0, 9937},
}}

\def\classNames{{"PERIOD","COMMA","QUESTION","EMPTY"}} 

\def\numClasses{4} 

\def\myScale{1.5} 
\begin{tikzpicture}[
    scale = \myScale,
    ]

\tikzset{vertical label/.style={rotate=90,anchor=east}}   
\tikzset{diagonal label/.style={rotate=45,anchor=north east}}

    \node [anchor=east] at (0.4,-1) {PERIOD}; 
    \node [anchor=east] at (0.4,-2) {COMMA}; 
    \node [anchor=east] at (0.4,-3) {QUESTION}; 
    \node [anchor=east] at (0.4,-4) {EMPTY};

    \node [rotate=90] at (1, -5.2) {PERIOD};
    \node [rotate=90] at (2, -5.235) {COMMA};
    \node [rotate=90] at (3, -5.36) {QUESTION};
    \node [rotate=90] at (4, -5.15) {EMPTY};

\foreach \y in {1,...,\numClasses} 
{
    \foreach \x in {1,...,\numClasses}  
    {
    \def\totSamples{0}
    \foreach \ll in {1,...,\numClasses}
    {
        \pgfmathparse{\myConfMat[\ll-1][\x-1]}   
        \xdef\totSamples{\totSamples+\pgfmathresult} 
    }
    \pgfmathparse{\totSamples} \xdef\totSamples{\pgfmathresult}  
    
    \begin{scope}[shift={(\x,-\y)}]
        \def\mVal{\myConfMat[\y-1][\x-1]} 
        \pgfmathtruncatemacro{\r}{\mVal}   %
        \pgfmathtruncatemacro{\p}{round(\r/\totSamples*100)}
        \coordinate (C) at (0,0);
        \ifthenelse{\y=\x}{\def\txtcol{white}}{\def\txtcol{black}}
        \node[
            draw,                 
            text=\txtcol,         
            align=center,         
            fill=black!\p,        
            minimum size=\myScale*10mm,    
            inner sep=0,          
            ] (C) {\r\\\p\%};     
    \end{scope}
    }
}
\coordinate (yaxis) at (-1.5,0.4-\numClasses/2);  
\coordinate (xaxis) at (0.5+\numClasses/2, -\numClasses-2.25); 
\node [vertical label] at (yaxis) {Predicted Class};
\node []               at (xaxis) {True Class};
\end{tikzpicture}

\label{confusion_human}
\caption{Confusion matrix for human predictions.}
\end{figure}

\subsection{Sample texts}\label{sec:sample_texts}
To investigate the supposed errors made by the human participants, some samples are presented along with the original source material that serve as the "correct" answer.

\subsubsection{Sample 1}\label{sec:halland}
This example, shown in figure \ref{example:landsbygd}, illustrates how the human participants are not intuitively wrong, even though they score low on the test. The participant's punctuation leaves out the first comma for a modern and less formal style without a subordinate clause and uses the so called "Oxford comma" after enumerating three (or more) items.

\noindent\begin{figure}
\fbox{\begin{minipage}{.47\textwidth}
Landsbygdsenheten har, utöver insatser inom jordbrukarstöd, landsbygdsprogram och rådgivning, fortsatt samverkan kring ett regionalt livsmedelsprogram.
\end{minipage}}\hfill\fbox{\begin{minipage}{.47\textwidth}
landsbygdsenheten har utöver insatser inom jordbrukarstöd, landsbygdsprogram, och rådgivning fortsatt samverkan kring ett regionalt livsmedelsprogram.
\end{minipage}}
\label{example:landsbygd}
\caption{Original punctuation (left) and human participant punctuation (right).}
\end{figure}

\subsubsection{Sample 2}\label{sec:gävleborg}
Shown in figure \ref{example:gävle} is a long-winded sentence with plenty of commas. It is a list of people and their profession, company and location. The syntax in the original text is "[name], [profession] at [employer], [location]" except for the last person, Inger Källgren Sawela, whose profession is not preceeded by a comma. The human participant has instead opted for the syntax "[name], [profession] at [employer] [location]". This is common syntax for companies, e.g. "McDonalds Leicester Square" and "Nintendo France" being the actual names of the company branches in their respective locations. Socialdemokraterna Gävleborg, for example, is the name on Google Maps, not "Socialdemokraterna, Gävleborg". Furthermore, the participant has merged the second sentence into the first by wrongly ascribing "länstyrelsens insynsråd" to be a part of Inger Källgren Sawela's employer. This is grammatically incorrect, as it should have been "Gävle länstyrel\textbf{ses}" if that were the case. This is also likely due to not observing the entire sentence, where "Länsstyrelsens insynsråd" is brought up in the start. This gives a clear signal that the second mention is from another sentence. Once again, the Oxford comma is used before the last person's details. These issues result in a total of eight errors (16 if we count "EMPTY").

The sentence has also been truncated due to the 650 word limit imposed during the test document generation. In this sample, the participant has correctly punctuated the start of their test despite the missing information. This gives a small piece of evidence that the human participants can handle the truncated sentences well and that they do not introduce a significant amount of errors.

\noindent\begin{figure}
\fbox{\begin{minipage}{.47\textwidth}
\textit{Länsstyrelsens insynsråd har under året, förutom av landshövdingen, bestått av följande sju ledamöter; Hanna Bruce,  VD för Växbo Lin AB, Växbo, Gabriella Persson Turdell, chef för Arbetsförmedlingen i norra Hälsingland, Raimo Pärssinen, riksdagsledamot för Socialdemokraterna, Gävle, Mehrdad Nesai, VD} för Glasakuten, Gävle och Lotta Zetterlund, ombudsman för LRF, Storvik, samt  Maj-Britt Johansson, tidigare rektor för Högskolan i Gävle och Bengt Kjellson, tidigare generaldirektör för Lantmäteriet, Gävle och Inger Källgren Sawela kommunalråd i Gävle. Länsstyrelsens insynsråd träffades fyra gånger under 2018, tre gånger för sammanträden och en gång för en studieresa till Gävlekusten. 
\end{minipage}}\hfill\fbox{\begin{minipage}{.47\textwidth}
\textit{Länsstyrelsens insynsråd har under året, förutom av landshövdingen, bestått av följande sju ledamöter; Hanna Bruce,  VD för Växbo Lin AB, Växbo, Gabriella Persson Turdell, chef för Arbetsförmedlingen i norra Hälsingland, Raimo Pärssinen, riksdagsledamot för Socialdemokraterna, Gävle, Mehrdad Nesai, VD} för glasakuten gävle och lotta zetterlund, ombudsman för lrf storvik, samt maj-britt johansson, tidigare rektor för högskolan i gävle, och bengt kjellson, tidigare generaldirektör för lantmäteriet gävle, och inger källgren sawela, kommunalråd i gävle länsstyrelsens insynsråd, träffades fyra gånger under 2018, tre gånger för sammanträden och en gång för en studieresa till gävlekusten.
\end{minipage}}
\label{example:gävle}
\caption{Original punctuation (left) and human participant punctuation (right). The italicized text was not seen by the participant as it belonged the the previous participant's test.}
\end{figure}

\subsubsection{Sample 3}\label{sec:livsmedelsverket}
This example highlights the heading problem. With varying frequency, the text includes headings with subject changes, as seen in figure \ref{example:livsmedelsverket} with "\#Bra matvanor". Since the hash symbol ("\#") and newlines are removed during preprocessing, these are harder to catch than normal. For this reason, they are often incorporated into surrounding sentences, as seen with "... och lär av andra bra matvanor." (eng. "and learn from \textit{other good eating habits}.") whereas the original sentence is "... och lär av andra." (eng. "and learn from \textit{others}."). This inter-weaving of subjects happen almost every time a new header appears.

There are some examples of grammatically incorrect punctuation from the participant. When enumerating three items, the participant has erroneously excluded a comma between the first and second item, e.g. in "... bly, kvicksilver och kadmium ..." (eng. "lead, mercury and cadmium"). This happened twice but in two other instances, the comma was included. This is therefore likely an oversight.

\noindent\begin{figure}
\fbox{\begin{minipage}{.47\textwidth}
en mycket låg andel överskridanden av gränsvärden, men det finns utmaningar. Intaget av de oönskade ämnena bly, kvicksilver och kadmium har inte minskat. [...] Förmågan att upptäcka fusk har stärkts genom en ökad samverkan med andra myndigheter och genom att Livsmedelsverket är aktiva i europeiska nätverk och lär av andra. 
\newline
\#Bra matvanor
\newline
Bra matvanor betyder hållbara matvanor [...]
\end{minipage}}\hfill\fbox{\begin{minipage}{.47\textwidth}
en mycket låg andel överskridanden av gränsvärden men det finns utmaningar, intaget av de oönskade ämnena bly kvicksilver och kadmium har inte minskat. [...] förmågan att upptäcka fusk har stärkts genom en ökad samverkan med andra myndigheter och genom att livsmedelsverket är aktiva i europeiska nätverk och lär av andra bra matvanor. bra matvanor betyder hållbara matvanor [...]
\end{minipage}}
\label{example:livsmedelsverket}
\caption{Original punctuation (left) and human participant punctuation (right).}
\end{figure}

\subsection{Variance}
As previously mentioned in section \ref{sec:method:human}, the participants for the experiment were a diverse group, representing ages 20 to 65, both men and women, been taught grammar in many different schools and decades, living in different geographical locations with distinct dialects and local linguistic discrepancies, education backgrounds in creative or quantitative fields, reading a lot or very little, writing in their spare time, having professional experience with correcting grammar.

It is not unthinkable that the data set could have some sort of bias. Perhaps some people are more inclined to use many commas where others do not or perhaps some write shorter sentences while others prefer longer. As seen in the example below in section \ref{sec:old_fiction}, old text (early 20th century) have a writing style rich with long-winded sentences with a lot of commas. Modern writing, seen in \ref{sec:recent_news}, tends to use more straightforward sentence structure with fewer subordinate clauses using commas. This can also differ between different writing contexts, such as a novel trying to invoke 
feelings and have a unique style
whereas a news article might favor less complexity to successfully and quickly convey information. Despite this, the standard deviation of overall F1 score of "PERIOD" and "COMMA" for the 16 test people is less than $0.1$. The "EMPTY" and "QUESTION" labels were left out since "EMPTY" is not included in the international reports and "QUESTION" since the F1 score for all but one particant was $0.0$ due to a lack of positive examples. Looking at the results, there seems to be no discernible pattern between test scores in any category and any personal attribute.

\section{International projects}
The international projects from Hungary and China have been trained on different data sets, since it is required that the data set is on the same language as it will then be used on. Otherwise, it is an M-BERT, which is not the goal of any of these projects. Thus, the comparison between the results is dubious, but there is no feasible alternative in the scope of this project. Translating the data sets and running the models on them is a substantial undertaking which includes resources which are not widely accessible. There are also, to the best of my knowledge, no other Swedish punctuation restoration projects to compare to.

\section{Sample outputs}
Two examples of outputs from prestoBERT have been included to visualize its power. These experiments have been performed with the fully-trained model. The reader is encouraged to try to read the input and then the output and judge for themselves if the output from prestoBERT is more readable and makes sense. Compare the output to the original excerpt: which one do you like the most?

\subsection{Recent KTH news}\label{sec:recent_news}
Recent newsletters from a government-owned institution is exactly what prestoBERT has been trained to restore punctuation for. As previously mentioned, the data set only contains publications from 2016 to 2019, so a recent article from KTH in April 2021 is guaranteed to be unseen by the model while still being similar to the training set. The excerpt, input and output are seen in figure \ref{kth1}, \ref{kth2} and \ref{kth3} respectively.

Just like anticipated in section \ref{sec:punctuation_restoration} with the ambiguous example "solen skiner idag är det varmt", the sequence "...  vidhåller 240 grader i två timmar efter oxideringsprocessen bryts ..." has two valid interpretations, assuming deep knowledge about natural science is ignored. The original text has "... vidhåller 240 grader \textit{i två timmar}. Efter oxideringsprocessen bryts ..." (eng. "... \textit{maintains 240 degrees for two hours}. After the oxidation process stops ...") but prestoBERT has the interpretation that "... vidhåller 240 grader. \textit{i två timmar efter} oxideringsprocessen bryts ..." (eng. "maintains 240 degrees. \textit{For two hours after that} the oxidation process stops ...". This is, although evidently false in the scientific sense, a valid interpretation that would be penalized during testing. This indicates that the performance of prestoBERT can in practice work better than the results imply, since the actual task is not to strictly "restore" but rather introduce valid punctuation.

Looking at the logits (likeliness scores for each class) in table \ref{table:recent_news}, we can see that the period placement is contentious. PrestoBERT has deemed "grader" a score of 3.3 for PERIOD and 2.6 for EMPTY and 2.8 and 3.6 respectively for "timmar". Having more than one positive logits for one token is unusual and having them so close in value is even more rare. This gives the impression that this decision "could go either way" and using a stochastic model to choose punctuation would yield the two alternatives almost equally often.

The only other error is two instances of closing commas in "... , en biprodukt från trämassa, ..." and arguably "... det fungerar lika bra, om inte bättre, än liknande hydrogelbaserade reningstekniker som används idag, ..." where the missed comma is after "bättre". 
However, the original "om inte bättre, än" is grammatically incorrect. When commas are used as a form of parentheses around an optional statement, the sentence has to work without the optional part. In the original text, reading without the "remark", we get "Det fungerar \textit{lika bra än} liknande ..." (eng. "It works \textit{as well than} similar ..."). There is no grammatically sound way to close the clause without adding another word ("som", eng. "as"), so prestoBERT's output is actually preferred over the original. Note that prestoBERT manages to close commas in "... , små kolbitar om 10 till 80 nanometer i diameter, ...".

Inspecting the logits again in table \ref{table:recent_news}, we observe that the first example ("trämassa"), the model in confident in the EMPTY label, although the possibility of a comma goes up in the sub-token "\#\#massa". This has no bearing on the final output since prestoBERT only predicts based on root token logits. In the second example ("bättre än"), the model misses both the grammatically incorrect and correct alternatives and is erroneously even more sure that there is no comma possibility than in the first example.

\begin{table}[]
\centering
\begin{tabular}{|lllll|}
\hline
\multicolumn{1}{|l|}{Token}     & \multicolumn{1}{l|}{PERIOD} & \multicolumn{1}{l|}{EMPTY} & \multicolumn{1}{l|}{COMMA} & QUESTION \\ \hline
\multicolumn{1}{|l|}{från}   & \multicolumn{1}{l|}{-4.3} & \multicolumn{1}{l|}{7.9} & \multicolumn{1}{l|}{0.4}  & -3.1 \\ \hline
\multicolumn{1}{|l|}{trä}    & \multicolumn{1}{l|}{-5.1} & \multicolumn{1}{l|}{7.7} & \multicolumn{1}{l|}{0.6}  & -2.8 \\ \hline
\multicolumn{1}{|l|}{\#\#massa} & \multicolumn{1}{l|}{-4.5}   & \multicolumn{1}{l|}{7.2}   & \multicolumn{1}{l|}{1.5}   & -2.9     \\ \hline
\multicolumn{1}{|l|}{till}   & \multicolumn{1}{l|}{-4.2} & \multicolumn{1}{l|}{7.2} & \multicolumn{1}{l|}{0.6}  & -2.7 \\ \hline
                             &                           & \multicolumn{1}{c}{...}  &                           &      \\ \hline
\multicolumn{1}{|l|}{grader} & \multicolumn{1}{l|}{3.3}  & \multicolumn{1}{l|}{2.6} & \multicolumn{1}{l|}{-2.6} & -3.5 \\ \hline
\multicolumn{1}{|l|}{i}      & \multicolumn{1}{l|}{-2.0} & \multicolumn{1}{l|}{8.5} & \multicolumn{1}{l|}{-3.3} & -2.8 \\ \hline
\multicolumn{1}{|l|}{två}    & \multicolumn{1}{l|}{-2.9} & \multicolumn{1}{l|}{9.0} & \multicolumn{1}{l|}{-3.1} & -2.7 \\ \hline
\multicolumn{1}{|l|}{timmar} & \multicolumn{1}{l|}{2.8}  & \multicolumn{1}{l|}{3.6} & \multicolumn{1}{l|}{-2.1} & -3.0 \\ \hline
\multicolumn{1}{|l|}{efter}  & \multicolumn{1}{l|}{-3.2} & \multicolumn{1}{l|}{9.1} & \multicolumn{1}{l|}{-2.5} & -3.2 \\ \hline
                             &                           & \multicolumn{1}{c}{...}  &                           &      \\ \hline
\multicolumn{1}{|l|}{inte}   & \multicolumn{1}{l|}{-3.5} & \multicolumn{1}{l|}{7.7} & \multicolumn{1}{l|}{-2.8} & -2.7 \\ \hline
\multicolumn{1}{|l|}{bättre}    & \multicolumn{1}{l|}{-2.6}   & \multicolumn{1}{l|}{7.1}   & \multicolumn{1}{l|}{-0.4}  & -3.6     \\ \hline
\multicolumn{1}{|l|}{än}     & \multicolumn{1}{l|}{-2.2} & \multicolumn{1}{l|}{7.7} & \multicolumn{1}{l|}{-1.4} & -3.5 \\ \hline
\end{tabular}
\label{table:recent_news}
\caption{Logits from the recent news example.}
\end{table}
 
\begin{figure}[!ht]
\centering
\fbox{\begin{minipage}{30em}
    KTH-forskarna har tagit fram en ny metod för att förkolna biomassan vid betydligt lägre temperaturer. De omvandlar lignosulfonat, en biprodukt från trämassa, till kolflingor genom att värma upp det i vatten i en mikrovågsugn. Vattnet hettas upp till 240 grader Celsius och vidhåller 240 grader i två timmar. Efter oxideringsprocessen bryts kolflingorna upp i små beståndsdelar, små kolbitar om 10 till 80 nanometer i diameter, som sedan blandas med cellulosagummit och behandla med UV-ljus för att forma hydrogelen. Detta är en enkel, hållbar lösning. Det fungerar lika bra, om inte bättre, än liknande hydrogelbaserade reningstekniker som används idag, säger Minna Hakkarainen.
\end{minipage}}
\label{kth1}
\caption{Excerpt from "Reningsteknik tillverkad i mikrovågsugn" from KTH news, published 23/4 2021 \cite{reningsteknik}.}
\end{figure}

\begin{figure}[!ht]
\centering
\fbox{\begin{minipage}{30em}
    kth-forskarna har tagit fram en ny metod för att förkolna biomassan vid betydligt lägre temperaturer de omvandlar lignosulfonat en biprodukt från trämassa till kolflingor genom att värma upp det i vatten i en mikrovågsugn vattnet hettas upp till 240 grader celsius och vidhåller 240 grader i två timmar efter oxideringsprocessen bryts kolflingorna upp i små beståndsdelar små kolbitar om 10 till 80 nanometer i diameter som sedan blandas med cellulosagummit och behandla med uv-ljus för att forma hydrogelen detta är en enkel hållbar lösning det fungerar lika bra om inte bättre än liknande hydrogelbaserade reningstekniker som används idag säger minna hakkarainen
\end{minipage}}
\label{kth2}
\caption{prestoBERT input (recent KTH news).}
\end{figure}

\begin{figure}[!ht]
\centering
\fbox{\begin{minipage}{30em}
    kth-forskarna har tagit fram en ny metod för att förkolna biomassan vid betydligt lägre temperaturer. de omvandlar lignosulfonat, en biprodukt från trämassa till kolflingor genom att värma upp det i vatten i en mikrovågsugn. vattnet hettas upp till 240 grader celsius och vidhåller 240 grader. i två timmar efter oxideringsprocessen bryts kolflingorna upp i små beståndsdelar, små kolbitar om 10 till 80 nanometer i diameter, som sedan blandas med cellulosagummit och behandla med uv-ljus för att forma hydrogelen. detta är en enkel, hållbar lösning. det fungerar lika bra, om inte bättre än liknande hydrogelbaserade reningstekniker som används idag, säger minna hakkarainen.
\end{minipage}}
\label{kth3}
\caption{prestoBERT output (recent KTH news).}
\end{figure}

\subsection{Old fiction}\label{sec:old_fiction}
KB-BERT has been trained on old fiction, but the fine-tuned punctuation restoration layer has not. Note that prestoBERT is not capable of adding any punctuation beyond periods, commas and question marks, such as quotation marks which is relevant in this scenario. Shown in figures \ref{selma1}, \ref{selma2} and \ref{selma3}, prestoBERT is somewhat successful in recreating the structure despite not recognizing all the obsolete spelling. For example, "himmelen" is in modern day most commonly spelt and taught as "himlen". In those cases, WordPiece embeddings can help, since "himmelen" is tokenized as "himmel \#\#en", which preserves the root "himmel" (eng. "sky") and thus its semantic meaning.

Investigating the errors, we can first observe that there are multiple false negative commas present, notably in the first sentence. It is important to point out that the original excerpt uses substantially more commas than modern Swedish tends to do. This can be a consequence of being written over 121 years ago and to some degree the nature of style in fiction, personal or not. As such, the output from prestoBERT is more akin to modern writing and grammatically sound. In the sequence "... fukt och fetma och han ...", the correct answer is to put a comma after "fetma" and "han", which prestoBERT fails to do. By inspecting the logits, we can observe something interesting, shown in table \ref{table:old_fiction}. The scores for "fukt" is an EMPTY prediction with high confidence, i.e. high positive value for one class and high negative values for the other classes. Looking at "fetma", the scores are significantly more uncertain, with scores generally close to zero and with two of them positive. In contrast, the logits for the first token, "jorden", is on the same level of confidence to both instances of the token "och", indicating that prestoBERT completely misses the possibility of a comma there.

It is disappointing that it misses the cue "kan" (eng. "could") in the question "kan det behövas ... i himmelen?". Notice how most tokens have low scores for question marks (around negative 3 and 4) whereas "himmel" has around half of that. The sub-token "\#\#en" has even more uncertainty with a lower score for PERIOD and even a postive score for QUESTION. This indicates that while incorrect, prestoBERT is closer to the correct answer than the output suggests at face value and is at least aware of the uncertainty.

\begin{table}[]
\centering
\begin{tabular}{|lllll|}
\hline
\multicolumn{1}{|l|}{Token}         & \multicolumn{1}{l|}{PERIOD} & \multicolumn{1}{l|}{EMPTY} & \multicolumn{1}{l|}{COMMA} & QUESTION \\ \hline
\multicolumn{1}{|l|}{jorden} & \multicolumn{1}{l|}{-4.1} & \multicolumn{1}{l|}{8.6}  & \multicolumn{1}{l|}{-1.7} & -3.0 \\ \hline
                             &                           & \multicolumn{1}{c}{...}   &                           &      \\ \hline
\multicolumn{1}{|l|}{pl (\#\#ogen)} & \multicolumn{1}{l|}{-4.2}   & \multicolumn{1}{l|}{7.0}   & \multicolumn{1}{l|}{0.4}   & -3.3     \\ \hline
                             &                           & \multicolumn{1}{c}{...}   &                           &      \\ \hline
\multicolumn{1}{|l|}{fukt}   & \multicolumn{1}{l|}{-3.2} & \multicolumn{1}{l|}{8.4}  & \multicolumn{1}{l|}{-1.8} & -3.1 \\ \hline
\multicolumn{1}{|l|}{och}    & \multicolumn{1}{l|}{-3.4} & \multicolumn{1}{l|}{8.9}  & \multicolumn{1}{l|}{-2.4} & -3.0 \\ \hline
\multicolumn{1}{|l|}{fetma}  & \multicolumn{1}{l|}{2.5}  & \multicolumn{1}{l|}{-0.1} & \multicolumn{1}{l|}{0.6}  & -3.7 \\ \hline
\multicolumn{1}{|l|}{och}    & \multicolumn{1}{l|}{-1.7} & \multicolumn{1}{l|}{6.5}  & \multicolumn{1}{l|}{-1.3} & -3.6 \\ \hline
\multicolumn{1}{|l|}{han}    & \multicolumn{1}{l|}{-4.0} & \multicolumn{1}{l|}{7.9}  & \multicolumn{1}{l|}{-0.6} & -3.4 \\ \hline
                             &                           & \multicolumn{1}{c}{...}   &                           &      \\ \hline
\multicolumn{1}{|l|}{leva}   & \multicolumn{1}{l|}{0.5}  & \multicolumn{1}{l|}{-1.1} & \multicolumn{1}{l|}{-1.2} & 1.4  \\ \hline
                             &                           & \multicolumn{1}{c}{...}   &                           &      \\ \hline
\multicolumn{1}{|l|}{himmel} & \multicolumn{1}{l|}{5.3}  & \multicolumn{1}{l|}{-1.6} & \multicolumn{1}{l|}{-2.4} & -1.7 \\ \hline
\multicolumn{1}{|l|}{\#\#en} & \multicolumn{1}{l|}{4.5}  & \multicolumn{1}{l|}{-3.2} & \multicolumn{1}{l|}{-3.2} & 1.1  \\ \hline
\end{tabular}
\label{table:old_fiction}
\caption{Logits from the old fiction example.}
\end{table}

\begin{figure}[!ht]
\centering
\fbox{\begin{minipage}{30em}
    Jorden, som vändes upp av plogen, låg svartbrun och sken av fukt och fetma, och han, som plöjde, gick och gladde sig åt att snart få så råg där. Han tänkte för sig själv: "Hur kan det komma sig, att jag stundom gör mig så svåra bekymmer och tycker, att det är tungt att leva? Kan det behövas något mer än solsken och vackert väder för att vara säll som ett Guds barn i himmelen?"
\end{minipage}}
\label{selma1}
\caption{Excerpt from "Jerusalem" by Selma Lagerlöf, 1901. \cite{selma}}
\end{figure}

\begin{figure}[!ht]
\centering
\fbox{\begin{minipage}{30em}
    jorden som vändes upp av plogen låg svartbrun och sken av fukt och fetma och han som plöjde gick och gladde sig åt att snart få så råg där han tänkte för sig själv : hur kan det komma sig att jag stundom gör mig så svåra bekymmer och tycker att det är tungt att leva kan det behövas något mer än solsken och vackert väder för att vara säll som ett guds barn i himmelen
\end{minipage}}
\label{selma2}
\caption{prestoBERT input (old fiction).}
\end{figure}

\begin{figure}[!ht]
\centering
\fbox{\begin{minipage}{30em}
    jorden som vändes upp av plogen låg svartbrun och sken av fukt och fetma. och han som plöjde gick och gladde sig åt att snart få så råg där. han tänkte för sig själv: hur kan det komma sig att jag stundom gör mig så svåra bekymmer och tycker att det är tungt att leva? kan det behövas något mer än solsken och vackert väder för att vara säll som ett guds barn i himmelen.
\end{minipage}}
\label{selma3}
\caption{prestoBERT output (old fiction).}
\end{figure}

\section{Attention visualization}
\begin{figure}[!ht]
  \begin{center}
    \includegraphics[width=0.9\textwidth]{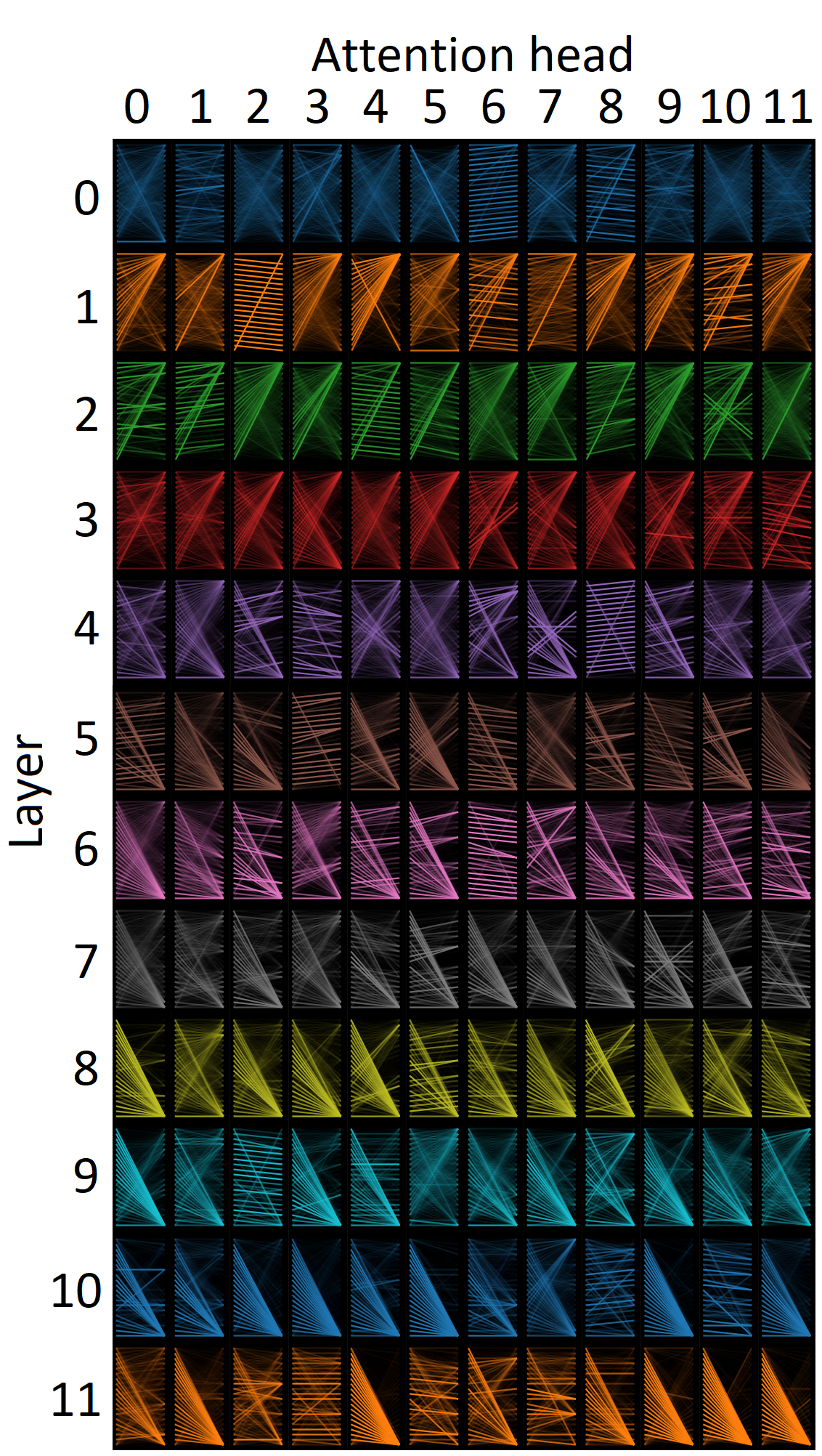}
  \end{center}
  \caption{Attention heads visualized using BertViz model view \cite{bertviz}. Each cell shows how words are related through attention. These are difficult to interpret, as many instances show large amounts of attention between the SEP or CLS token going to every other token with roughly equal strength, rather than illustrating word relationships. See figure \ref{L4H2_black} for a zoom-in on layer 4, head 2.}
  \label{attention_layers}
\end{figure}

\begin{figure}[!ht]
  \begin{center}
    \includegraphics[scale=0.9]{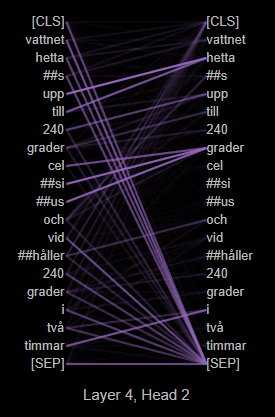}
  \end{center}
  \caption{Layer 4, attention head 2 visualized using BertViz model view \cite{bertviz}. The attention between "grader" (eng. "degree") and celsius makes intuitive sense as the word "grader" has multiple different units of measurement, e.g. Fahrenheit, pi, 360 degree circle parts and color hues.}
  \label{L4H2_black}
\end{figure}

The attention heads in different layers can be visualized in color, like in section \ref{sec:attention}. Every attention head present in prestoBERT is visualized in figure \ref{attention_layers} for the sentence "vattnet hettas upp till 240 grader celsius och vidhåller 240 grader i två timmar.". This is known as a model view and shows a coarse shape of the attention pattern, i.e. a quick overview. This can be used to identify particularly interesting attention heads for further investigation \cite{bertviz}. It is possible to zoom in on individual images for more detail, exemplified in figure \ref{L4H2_black}. In these figures, the left hand side are query vectors for each token and the right hand side are their key vectors. The stronger the line between a query and key vector, the higher their product is.

\cleardoublepage
\chapter{Discussion}\label{ch:discussion}

\section{prestoBERT}\label{sec:disc:prestoBERT}
The sample outputs highlight the ambiguity deeply rooted in punctuation restoration. The overarching goal is to develop a model that can correctly punctuate text. It is clear that the underlying formulation of the desired result, to return exactly the same sentence as the original, works well enough, but is not ideal. There is a sort of disconnect, since the training phase punishes behaviour that is positive, e.g. outputting a different but still valid interpretation of the source material. The consequence of this is unclear.
It likely affects the F1-score negatively, but the disparity between prestoBERT and Hubert is most likely due to the significantly smaller data set used when training the former. This seems to not have affected the COMMA and PERIOD results much, but looking at QUESTION that is almost unusable and unstable, one has to wonder how much the former two could be improved with additional data. While the QUESTION class had only 2.1\% of the cases in the Hungarian data set, COMMA and PERIOD did not have much more at 7.9\% and 21.1\% respectively. The performance drop for the QUESTION class implies that data is important, but the relatively high F1-score for the other two suggests that there is either a diminishing return of data, i.e. that somewhere around 10,911 commas is enough to learn most weights, or that the ceiling for PERIOD and COMMA prediction is higher and data set size is a bottle neck.

There is no solid evidence either way. On the one hand, Hubert and prestoBERT performed better than the Chinese BERT-BLSTM-CRF despite the latter having roughly three times and six times the data compared to the former two respectively. On the other hand, the instability in prestoBERT's QUESTION class due to a lack of data is evident. Regardless, it is evident that even a small data set can result in decent model, since prestoBERT has an overall F1-score in between two state-of-the-art models.

Fortunately, in the context of punctuation restoration, generating more data is straightforward. There is no particular or custom text required for this problem, only regular, punctuated text, like the ones used during the pre-training of KB-BERT. There is a concern that fine-tuning the output layer on the same data set as the BERT model was pre-trained on might allow the model to overfit by remembering punctuation from pre-training. Investigating this could be important for future work, since reusing the same data makes data usage highly efficient, whereas the opposite would require a pre-train/fine-tune split in data. For example, since KB-BERT used the entirety of Swedish Wikipedia, arguably the most easily accessible data set online, during pre-training, it would be significantly more difficult for the fine-tuning project to find new data. It is worth mentioning that the data used in the fine-tuning project was likely not used in the pre-training of KB-BERT.

The difference in performance could also be due to differences in architecture. The Chinese model is made more complex by adding additional neural networks on top of BERT, while the Hungarian model uses a sliding window. This is also hard to evalutate, since the most complex (Chinese) performs worse than the less complex (Hungarian), but which still beats the least complex (Swedish). One noteworthy difference is that prestoBERT uses a cased version of KB-BERT, since an uncased version does not exist at the time of writing, but the Hungarian project showed that for handling uncased data, an uncased model is superior. It would therefore likely be a significant improvement to this endeavour and similar projects if the National Library developed an uncased KB-BERT model.

Above have been reasons why prestoBERT is worse than its international counterparts, but what are the reasons that it in some cases performed better? It seems incredible that prestoBERT sometimes beats more complex models with substantially more data, developed by large teams of experienced scientists. In the case of QUESTION, it is likely due to a small sample size during testing, since the performance in the original test (presented in \ref{table:results}) is unusually good while in the confusion matrix it is unusually poor. However, in terms of PERIOD, it probably had sufficient samples. We might instead look at the influence of batching bias brought up in section \ref{sec:feeding_into_BERT}. Even discounting the affected instances of PERIOD, knowing a PERIOD for certain is likely to help piece together the rest of the sentence. Punctuation restoration is somewhat akin to solving a jigsaw puzzle or Sodoku; the more pieces that are in their correct places, the easier it is to lay the next piece. For example, if we know the punctuation at position $i$ is a period, then we can most likely disregard the possibility of the punctuation at position $i-1$ being a question mark, since one-word sentences are exceedingly rare. If the word a position $i$ is a name of someone being addressed in conversation, the the punctuation at position $i-1$ could more likely be a comma (e.g. "Hello there, Clarence".). The bias introduced by the chosen batching method could have ripple effects far beyond a few trivial PERIOD predictions. The surprising performance of prestoBERT could imply that the bias is highly influential.

The alternative is not palatable, either, since it would involve sending in incomplete and incoherent sentences as input. We have seen how the human participants struggled to make sense of partially masked sentences in section \ref{sec:sample_texts}. Since BERT only supports around 7 sentences per batch, allowing the first and last sentence to be potentially unsolvable due to lack of information risks wasting precious data and teaching the model weird strategies that are unusable in the real-world scenario where incomplete gibberish is not input.

The choice of which token prediction to translate into the predicted label for the whole word is not obvious. The only examples used in practice seem to be either first or last token prediction. In some cases, such as "tyrannosaurus", it makes sense to base the prediction on "\#\#saurus" rather than "ty", since "saurus" contains more information than just "ty". In other cases, such as "hettas", it, conversely, makes more sense to predict on "hetta" rather than "\#\#s". Which part of the word is larger might depend heavily upon the language. If the language is rich in prefixes, the prefix-parts ("ty") can be generic and short whereas if the language uses more suffixes, the suffix ("s") can be the generic part.

\subsection{Attention}
\begin{figure}[!ht]
  \begin{center}
    \includegraphics[scale=0.9]{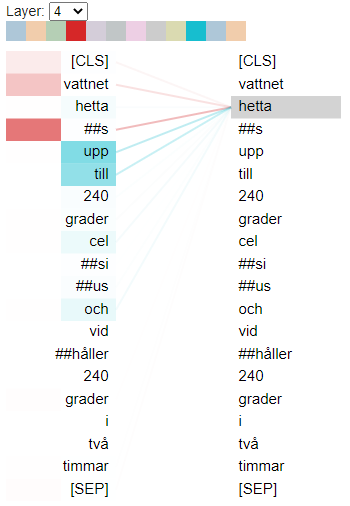}
  \end{center}
  \caption{Layer 4, attention heads 2 and 3 regarding the token "hetta". Visualized using BertViz head view \cite{bertviz}.}
  \label{L4H2_white}
\end{figure}

Visualizing the attention heads of a model is a way of increasing interpretability, i.e. understanding the model. However, looking at figure \ref{attention_layers}, most visuals show the "[CLS]" and "[SEP]" connecting to every token and not much else. There are also multiple diagonally striped images where each word attends to the one before or after it, which also does not offer much in terms of interpretability. However, a few of the images, such as figure \ref{L4H2_black}, hint at intuitive deductions similar to those exemplified in section \ref{sec:attention}. In figure \ref{L4H2_white}, two different attention heads in the same layer are visualized in the regards to the root token "hetta". The original word is "hettas" (eng. "is heated"), spliced into the tokens "hetta" and "\#\#s", where "hetta" is either a verb or noun depending on the context. The blue attention head (head 2) is attending to the tokens "upp till" (eng. "heated up to") which makes it clear that "hetta" is used in its verb form. The red attention head (head 3) instead attends to the sub-token "\#\#s" and "vattnet" (eng. "the water"), which suggest that it pieces together what is being heated.

The visualizations offered by BertViz by and large do not tell us how the model punctuates. It gives us an inkling of what words prestoBERT thinks are related. How exactly it determines that a word should be succeeded by a comma remains unclear. Looking at the logits as a measure of uncertainty might offer more insight in understanding prestoBERT.

\section{Human evaluation}
The small sample size compared to the test set of the model is non-ideal. However, given how the performance has low variance, the sampled population might be sufficiently indicative of the entire population, despite the small sample size.

The tests were not equal. The data was partitioned without overlap to increase coverage and it is not unlikely that some tests were more difficult than others, at least in terms of periods and commas. For example, one test contained 49 commas, meaning that almost every 13th word preceeded a comma, whereas another one contained only 7. This could have had a significant effect on the test results, but it is impossible to tell from this experiment alone.

It is somewhat counter-intuitive that human labels would get so low scores. After all, they constructed sentences that they themselves felt were legible and grammatically correct, yet they scored substantially worse compared to neural networks. How can there be such a big disconnect? It is likely that ambiguity is the main cause of low test results. As touched upon several times in this report, sentences can have multiple different interpretations, but only one is considered correct in the experiment. We have seen grammatically incorrect punctuation from both the participants and the source material, but also how different interpretations can all make sense. It seems that this test is not a suitable method to test general punctuation skills. However, there is, to also evidence of mistakes being made, albeit to a lesser extent than ambiguity. This is might be because of a lack of focus. Since almost all participants, unprompted, expressed that the task was fun and more difficult than they anticipated, it is likely that this lack of focus stems from tiredness rather than an absence of enthusiasm. Indeed, the source material is factoids about "current" events from the past, relevant only to those who live in certain counties and regions in Sweden a few years ago. This does not make for a page-turner and it likely strains focus and attention span of the participant to take this text to heart, especially given that they are looking at a "wall of text", devoid of any formatting that improves legibility, such as paragraph division, punctuation and capital letters.

The tests often had their first and/or last sentence truncated to accommodate the 650 word limit. As seen in sample 2 in section \ref{sec:gävleborg}, it is possible for the participants to pick up a partial sentence and correctly punctuate. This might not always be the case, though, and it is possible that the truncation introduced unnecessary errors in a manner that prestoBERT did not have to deal with. The alternative of keeping sentences intact and aiming for roughly 650 words per test would have eliminated this transitional error where tests start and end. It is unlikely that this would have had a significant impact on the test results, since the errors observed in section \ref{sec:sample_texts} are primarily due to alternative, yet valid, interpretations, not misunderstood context.

\section{Ethical analysis}
\subsection{Model bias}
Bias in machine learning models can arise due to bias in the data, since the model is programmed to pick up on patterns in the training set. Consequences range from unfair treatment to perpetuated stereotyping. In this project, this would likely incarnate as the punctuation having a certain "tone". This could be that the is data masculine, rural, technical, unusually complex or any number of biases. There is a possibility that the data used in this project is biased in some way, for example that it is primarily written by a sub-category of the Swedish population demographic. However, the results of the human evaluation indicate that there is no set of attributes particularly "in tune" with the data set, implying that there is no significant bias.

\subsection{Ecological sustainability}\label{sec:eco_sus}
Being a digital product, the model does not have much direct ecological impact beyond electricity consumption. However, it has to run on hardware, and particularly the training phase makes use of advanced electronics such as GPUs. These electronics impact the environment after their lifespan is over, since their intricate architecture make them difficult to recycle. However, the training can be done through cloud computing, which makes use of a shared set of GPUs. This means that the GPUs are used to their fullest extent, since many users make use of them whenever the units are available, rather than everyone having to own their own substantial capacity of hardware and having it age during inactivity. By cloud computing and only having to train once, the ecological impact of hardware is minimized.

The electrical consumption is substantial during pre-training of the underlying BERT model. However, since this, too, only has to be done once before it can be shared with any Swedish-based project for any NLP problem, this demand is also minimized. Additionally, the alternative, performance-wise, is to have human employees do the same work at what has been established in section \ref{res:prestoBERT} 6300 times slower rate. This costs electricity in anything from keeping the monitor running to wasted man-hours that could do other jobs.

\subsection{Social sustainability}
As with most generative models, prestoBERT could be used to automatize text-based scams, such as spam emails. One way to recognize spam is through unnatural language, which includes punctuation. Using a model trained to simulate human punctuation patterns could make it more difficult to discern whether text is written by a malicious bot or a person. The implication is that if it is written by a person, it is more likely that it is non-malicious since sending hand-written spams to thousands of people is largely unfeasible compared to automating the process.

The large-scale text generation model GPT-2 developed by OpenAI raised similar concerns. Not only did it generate punctuation, but entire text sequences. The potential risk for phishing and spam emails was frequently brought up, but ultimately nothing seems to have come from it. My model is substantially lesser scale than the GPT-2 and larger projects with the exact same goal of punctuation restoration have, to the best of my knowledge, not resulted in any spam coup d'etat.

To the best of my knowledge, there seems to be no open source model for automating punctuation in Swedish. The technology exists, since Google offers multiple services with speech-to-text features. This project (and KB-BERT) gives the opportunity for anyone with interested in the model by hosting it publicly available on HuggingFace's servers.

\subsection{Economical sustainability}
While the model does not produce infallible output and thus still needs a degree of oversight by humans, it is still, as mentioned earlier in section \ref{sec:eco_sus}, 6300 times as fast. This carries economic implications. Not only is it automated, which excludes the hourly wage of a person doing the same job, the increased speed makes the entire "assembly line" more efficient. Whatever the punctuated text is supposed to be used for, the next step in the process can start earlier, i.e. closer in time to when the unpunctuated text was handed off to be punctuated, instead of having to wait.

\cleardoublepage
\chapter{Conclusions and Future work}

\section{Conclusions}\label{sec:conclusions}
In conclusion, it seems that a fine-tuned BERT for punctuation restoration works well in Swedish in the same manner as other non-English languages. The Swedish language does not seem particularly difficult in terms of punctuation restoration, as the more complex, Chinese model struggles more than the simple architecture of prestoBERT despite having more data. Given how successful prestoBERT is with its limited data set and bare-bones architecture, it suggests that each language should find reason to develop their own BERT model rather than rely on M-BERT, especially for a problem like punctuation restoration where more data is generally easy to generate.

It seems that prestoBERT beats human participants in the task with a substantial margin, but it turns out that the task, as defined for this project, does not accurately reflect the goal of "figuring out suitable punctuation for a piece of text". Rather, it ended up being closer to "mimic the exact style of punctuation used in this data set". This is clearly not ideal, since many correct punctuations are deemed incorrect even though they are grammatically sound. As such, there is a form of information loss in the model's learning phase. It still works well, since it produces satisfactory sentences, but one has to wonder how much a more suitable formulation of the task could improve performance.

\section{Limitations}\label{sec:limitations}
The limiting factors for this project have been:
\begin{itemize}
    \item small data set (by choice),
    \item batch size limit,
    \item lack of an uncased KB-BERT and
    \item simple architecture (by choice).
\end{itemize}

\section{Future work}\label{sec:futureWork}
For future projects, attending to some of the limitations should likely result in immediate performance boosts.

The small data set is remedied by pulling more written text from the internet. Using literature that has been proofread is safe while using internet forums etc. where grammar is less strict could prove detrimental.

The batch size limit forces the algorithm to feed rather small subsets of the text one at a time with no cross-referencing in between. This is problematic since it either has to split by sentence, which introduces bias in exploiting the known period in the end of the batch, or by words, which makes many sentences nonsensical. Technical advances is needed to expand the limit on the number of tokens allowed per batch, but a clever algorithm on the software end could minimize this problem.

Using an uncased pre-trained BERT rather than a cased one would be ideal, since cased letter do not appear in the data for this problem. This is unfeasible for most entities to develop themselves, but there is hopefully incentive for the National Library to continue their work and develop an uncased KB-BERT.

The architecture used for prestoBERT is the simplest possible: pre-trained BERT and a fine-tuned layer stacked on top of it. Other languages have introduced even more neural networks on top or modified the attention mechanism with a sliding window offering more predictions and taking their average instead of relying on a single prediction.

Lastly, experimenting with which WordPiece token(s) to base predictions on result in finding a better approach than the root token prediction used in this report.

\cleardoublepage
\bibliographystyle{myIEEEtran}
\renewcommand{\bibname}{References}
\addcontentsline{toc}{chapter}{References}
\bibliography{thesis}

\cleardoublepage
\appendix
\renewcommand{\chaptermark}[1]{\markboth{Appendix \thechapter\relax:\thinspace\relax#1}{}}
\chapter{Test instructions}
\label{appendix_instructions}
\includepdf[pages=-]{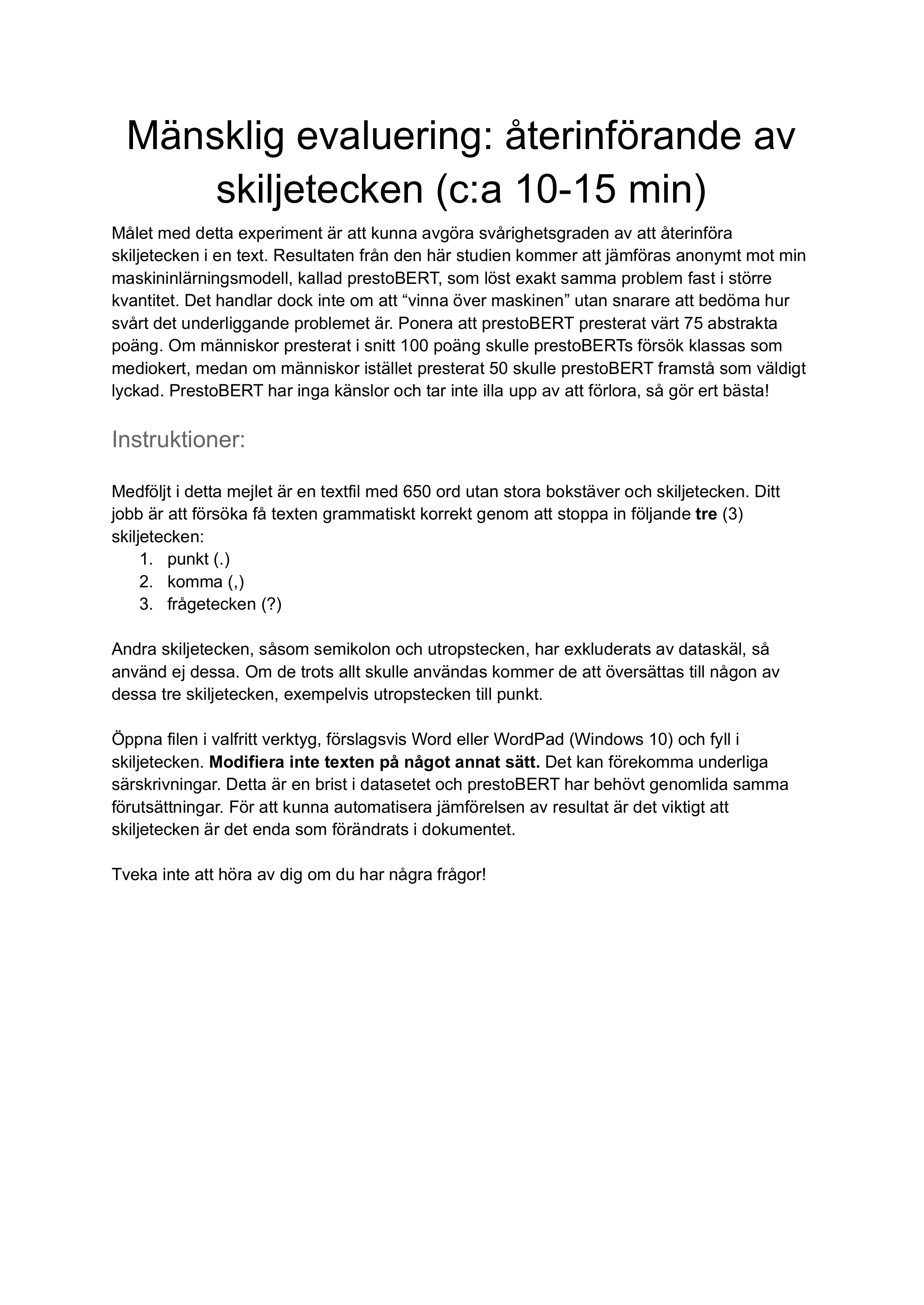}

\chapter{Sample test from human evaluation}
\label{appendix_sample_test}
inkluderas i disciplinära ärenden och vilka faktiska konsekvenser det kan medföra inte minst utifrån internationella studenters perspektiv som många gånger kommer från andra akademiska bakgrunder vi är positiva till universitetets egna satsningar för att informera berörda aktörer inom ämnet bland annat genom ett webinarie för medarbetare samt en uppdaterad frågedel på hemsidan arbetet med samverkansgruppen för strategiskt partnerskap mellan linn studenterna linn universitetet samt lärosätets två huvudsakliga kommuner har varit relevant för kårens studentpolitiska frågor och vi ser fram emot gruppens fortsatta arbete för kompetensförsörjning genom denna samverkansgrupps arbete ser vi att fler studenter kommer att välja växjö och kalmar som hemort efter avslutade studier under hösten har linn studenterna bjudits in till planeringsarbetet för utvecklingen av campus samt växjö som universitetsstad detta som en del i arbetet att förbättra studenters utbildnings - såväl som studiesociala miljö vi ser flera möjligheter att under arbetets gång kunna inkludera universitetets studenter i processen de 248 nya studentbostäder som tillkommit i växjö inför hösten 2018 har bidragit till att fler nya studenter har fått bostad jämfört med hösten 2017 vi ser däremot att behovet av bostäder i nuläget fortfarande är större än tillgången denna faktor tillsammans med högt satta hyror är anledningarna till att växjö även detta år har rödlistats sveriges förenade studentkårers ( sfs ) årliga bostadsrapport då bostadsfrågan är viktig för våra medlemmar är detta ett område vi fokuserar på i vårt arbete vår förhoppning är att vi i olika samarbeten kommer kunna stabilisera nivån på hyrorna för att i sin tur kunna öka tillgängligheten för heltidsstudier vid linn universitetet 2018 ett år med nya utmaningar 2018 var ett år som präglades av förändringar och osäkerhet i sverige och runt om i världen för livsmedelsverket innebar det nya utmaningar det blev tydligt hur viktiga våra ansvarsområden mat och dricksvatten är 2018 blev året då frågan om klimat och extremväder blev konkret även i sverige sommaren var rekordvarm i så gott som hela landet med utbredd torka vattenbrist stora skogsbränder betes - och foderbrist algblomning och låga grundvattennivåer den varma sommaren och hur den påverkade vår vardag blev en väckarklocka för många : en föraning om vad vi kan vänta oss och vad vi måste förbereda oss på som ett resultat av klimatförändringar vårt sätt att producera och konsumera livsmedel påverkar miljön på flera sätt och det krävs stora förändringar för att minska livsmedelssektorns miljöbelastning att en tredjedel av all mat som produceras slängs någonstans på vägen till våra tallrikar är inte bara ett slöseri utan orsakar stora och onödiga miljöbelastningar under sommaren presenterade vi tillsammans med naturvårdsverket och jordbruksverket handlingsplanen fler gör mer för minskat matsvinn i sverige handlingsplanen omfattar hela livsmedelskedjan från jord till bord och innehåller 42 konkreta åtgärder måltider för miljö hälsa och jämlikhet att vår livsmedelskonsumtion påverkar både miljön och hälsan är ingen nyhet : inte heller att den mat som är bra för miljön ofta gynnar människors hälsa därför betonar vi vikten av hållbarhet utifrån både hälsa och miljö i de uppdaterade råden för måltiderna inom skolan respektive äldreomsorgen måltiderna inom den offentliga sektorn är ett viktigt verktyg för att uppnå både miljömål och jämlik hälsa resultaten från vår nationella matvaneundersökning riksmaten ungdom oroar ungdomars matvanor riskerar att leda till ohälsa särskilt bland socioekonomiskt svagare grupper det är en utmaning för oss och för hela samhället mat och vatten när det krisar de säkerhetspolitiska förändringarna i vår omvärld ställer nya krav på livsmedelsverket som bevakningsansvarig myndighet arbetet med att bidra till att bygga upp det civila försvaret har intensifierats under året vi behöver fortsätta att utveckla vår förmåga att värna livsmedelssäkerheten vid långvariga kriser och ta höjd för nya faror samtidigt ligger alltmer fokus på livsmedelsförsörjningsfrågor hur ska vi få mat på bordet om något stör det vanliga flödet av varor här kommer livsmedelsverket att ha en viktig roll tillsammans med en rad samhällsaktörer framöver att värna dricksvattnet vårt viktigaste livsmedel är ett av livsmedelsverkets uppdrag

\label{pg:lastPageofMainmatter}

\clearpage
\end{document}